\pdfoutput=1

\documentclass{article}
\usepackage{ijcai15}

\usepackage{times}
\usepackage{url}
\usepackage{graphicx}
\usepackage{subfigure}
\usepackage{float}
\usepackage{enumerate}
\usepackage{bbm}
\usepackage{amsmath}
\usepackage{amsfonts}
\usepackage{amsthm}
\usepackage{algorithm}
\usepackage{algorithmic}

\title{ A New Relaxation Approach to Normalized Hypergraph Cut}

\author{
	Cong Xie\\
	Department of Computer\\ Science and Engineering\\
	Shanghai Jiao Tong University \\
	800 Dong Chuan Road \\
	Shanghai, China 200240\\
	xcgoner1108@gmail.com
\And
	Wu-Jun Li\\
	National Key Laboratory for\\ Novel Software Technology\\
    Department of Computer\\ Science and Technology\\
	Nanjing University\\
    liwujun@nju.edu.cn 
\And
	Zhihua Zhang\\
    Department of Computer\\ Science and Engineering \\
    Shanghai Jiao Tong University \\
    800 Dong Chuan Road \\
    Shanghai, China 200240\\
    zhang-zh@cs.sjtu.edu.cn
}

\begin{document}

\def\Blue{\color{blue}}
\def\Purple{\color{purple}}

\def\A{{\bf A}}
\def\a{{\bf a}}
\def\B{{\bf B}}
\def\C{{\bf C}}
\def\c{{\bf c}}
\def\D{{\bf D}}
\def\d{{\bf d}}
\def\E{{\bf E}}
\def\F{{\bf F}}
\def\e{{\bf e}}
\def\f{{\bf f}}
\def\G{{\bf G}}
\def\H{{\bf H}}
\def\I{{\bf I}}
\def\J{{\bf J}}
\def\K{{\bf K}}
\def\L{{\bf L}}
\def\M{{\bf M}}
\def\m{{\bf m}}
\def\N{{\bf N}}
\def\n{{\bf n}}
\def\Q{{\bf Q}}
\def\q{{\bf q}}
\def\R{{\bf R}}
\def\S{{\bf S}}
\def\s{{\bf s}}
\def\T{{\bf T}}
\def\U{{\bf U}}
\def\u{{\bf u}}
\def\V{{\bf V}}
\def\v{{\bf v}}
\def\W{{\bf W}}
\def\w{{\bf w}}
\def\X{{\bf X}}
\def\x{{\bf x}}
\def\Y{{\bf Y}}
\def\y{{\bf y}}
\def\Z{{\bf Z}}
\def\z{{\bf z}}
\def\0{{\bf 0}}
\def\1{{\bf 1}}

\def\AM{{\mathcal A}}
\def\CM{{\mathcal C}}
\def\DM{{\mathcal D}}
\def\GM{{\mathcal G}}
\def\FM{{\mathcal F}}
\def\IM{{\mathcal I}}
\def\NM{{\mathcal N}}
\def\OM{{\mathcal O}}
\def\SM{{\mathcal S}}
\def\TM{{\mathcal T}}
\def\UM{{\mathcal U}}
\def\XM{{\mathcal X}}
\def\YM{{\mathcal Y}}
\def\RB{{\mathbb R}}

\def\TX{\tilde{\bf X}}
\def\tx{\tilde{\bf x}}
\def\ty{\tilde{\bf y}}
\def\TZ{\tilde{\bf Z}}
\def\tz{\tilde{\bf z}}
\def\hd{\hat{d}}
\def\HD{\hat{\bf D}}
\def\hx{\hat{\bf x}}
\def\TD{\tilde{\Delta}}

\def\alp{\mbox{\boldmath$\alpha$\unboldmath}}
\def\bet{\mbox{\boldmath$\beta$\unboldmath}}
\def\epsi{\mbox{\boldmath$\epsilon$\unboldmath}}
\def\etab{\mbox{\boldmath$\eta$\unboldmath}}
\def\ph{\mbox{\boldmath$\phi$\unboldmath}}
\def\pii{\mbox{\boldmath$\pi$\unboldmath}}
\def\Ph{\mbox{\boldmath$\Phi$\unboldmath}}
\def\Ps{\mbox{\boldmath$\Psi$\unboldmath}}
\def\tha{\mbox{\boldmath$\theta$\unboldmath}}
\def\Tha{\mbox{\boldmath$\Theta$\unboldmath}}
\def\muu{\mbox{\boldmath$\mu$\unboldmath}}
\def\Si{\mbox{\boldmath$\Sigma$\unboldmath}}
\def\si{\mbox{\boldmath$\sigma$\unboldmath}}
\def\Gam{\mbox{\boldmath$\Gamma$\unboldmath}}
\def\Lam{\mbox{\boldmath$\Lambda$\unboldmath}}
\def\De{\mbox{\boldmath$\Delta$\unboldmath}}
\def\Ome{\mbox{\boldmath$\Omega$\unboldmath}}
\def\TOme{\mbox{\boldmath$\hat{\Omega}$\unboldmath}}
\def\vps{\mbox{\boldmath$\varepsilon$\unboldmath}}
\newcommand{\ti}[1]{\tilde{#1}}
\def\Ncal{\mathcal{N}}
\def\argmax{\mathop{\rm argmax}}
\def\argmin{\mathop{\rm argmin}}
\providecommand{\abs}[1]{\lvert#1\rvert}
\providecommand{\norm}[2]{\lVert#1\rVert_{#2}}

\def\Zs{{\Z_{\mathrm{S}}}}
\def\Zl{{\Z_{\mathrm{L}}}}
\def\Yr{{\Y_{\mathrm{R}}}}
\def\Yg{{\Y_{\mathrm{G}}}}
\def\Yb{{\Y_{\mathrm{B}}}}
\def\Ar{{\A_{\mathrm{R}}}}
\def\Ag{{\A_{\mathrm{G}}}}
\def\Ab{{\A_{\mathrm{B}}}}
\def\As{{\A_{\mathrm{S}}}}
\def\Asr{{\A_{\mathrm{S}_{\mathrm{R}}}}}
\def\Asg{{\A_{\mathrm{S}_{\mathrm{G}}}}}
\def\Asb{{\A_{\mathrm{S}_{\mathrm{B}}}}}
\def\Or{{\Ome_{\mathrm{R}}}}
\def\Og{{\Ome_{\mathrm{G}}}}
\def\Ob{{\Ome_{\mathrm{B}}}}

\def\vec{\mathrm{vec}}
\def\fold{\mathrm{fold}}
\def\index{\mathrm{index}}
\def\sgn{\mathrm{sgn}}
\def\tr{\mathrm{tr}}
\def\rk{\mathrm{rank}}
\def\diag{\mathsf{diag}}
\def\const{\mathrm{Const}}
\def\dg{\mathsf{dg}}
\def\st{\mathsf{s.t.}}
\def\vect{\mathsf{vec}}
\def\MCAR{\mathrm{MCAR}}
\def\MSAR{\mathrm{MSAR}}
\def\etal{{\em et al.\/}\,}
\newcommand{\indep}{{\;\bot\!\!\!\!\!\!\bot\;}}

\def\Lsize{\hbox{\space \raise-2mm\hbox{$\textstyle \L \atop \scriptstyle {m\times 3n}$} \space}}
\def\Ssize{\hbox{\space \raise-2mm\hbox{$\textstyle \S \atop \scriptstyle {m\times 3n}$} \space}}
\def\Osize{\hbox{\space \raise-2mm\hbox{$\textstyle \Ome \atop \scriptstyle {m\times 3n}$} \space}}
\def\Tsize{\hbox{\space \raise-2mm\hbox{$\textstyle \T \atop \scriptstyle {3n\times n}$} \space}}
\def\Bsize{\hbox{\space \raise-2mm\hbox{$\textstyle \B \atop \scriptstyle {m\times n}$} \space}}

\newcommand{\twopartdef}[4]
{
	\left\{
		\begin{array}{ll}
			#1 & \mbox{if } #2 \\
			#3 & \mbox{if } #4
		\end{array}
	\right.
}

\newcommand{\tabincell}[2]{\begin{tabular}{@{}#1@{}}#2\end{tabular}}

\renewcommand{\algorithmicrequire}{\textbf{Input:}} 
\renewcommand{\algorithmicensure}{\textbf{Output:}} 

\maketitle


 \begin{abstract}
 Normalized graph cut~(NGC) has become a popular research topic due to 
 its wide applications in a large variety of areas like machine learning 
 and very large scale integration~(VLSI) circuit design. Most of 
 traditional NGC methods are based on pairwise 
 relationships~(similarities). However, in real-world applications  
 relationships among the vertices~(objects) may be more complex than 
 pairwise, which are typically represented as hyperedges in hypergraphs.  
 Thus, normalized hypergraph cut~(NHC) has attracted more and more 
 attention. Existing NHC methods cannot achieve satisfactory 
 performance in real applications. In this paper, we propose a novel 
 relaxation approach, which is called \emph{\mbox{relaxed 
 NHC}}~(\mbox{RNHC}), to solve the NHC problem.  Our model is defined as 
 an optimization problem  on the Stiefel manifold. To solve this problem, 
 we resort to the Cayley transformation to devise a feasible learning  
 algorithm.  Experimental results on a set of large hypergraph benchmarks 
 for clustering and partitioning in VLSI domain  show that \mbox{RNHC} 
 can outperform the state-of-the-art methods.
 \end{abstract}

\section{Introduction}

The goal of graph cut~(or called graph partitioning)~\cite{DBLP:journals/pami/WuL93} is to divide the vertices~(nodes) in a graph into several groups~(clusters),  making the number of edges across different clusters minimized while the number of edges within the clusters  maximized. Besides the goal which should be achieved in graph cut, normalized graph cut~(NGC) should also make the volumes of different clusters as balanced as possible by adopting some normalization techniques. In many real applications,  NGC has been proved to achieve better performance than unnormalized graph cut \cite{DBLP:journals/pami/ShiM00,ng2002spectral,gonzalez2012powergraph}. Thus, NGC is a popular theme due to its wide applications in a large variety of areas, including machine learning~\cite{ng2002spectral,xie2014distributed}, parallel and distributed computation~\cite{gonzalez2012powergraph,jain2013graphbuilder,chen2013powerlyra}, image segmentation~\cite{DBLP:journals/pami/ShiM00}, and social network analysis~\cite{DBLP:series/synthesis/2010Tang,DBLP:conf/ijcai/LiS11}, and so on. For example, graph-based clustering methods such as spectral clustering~\cite{ng2002spectral}  can be seen as NGC methods. In social network analysis, NGC has been widely used for community detection in social networks~(graphs).

Most of traditional NGC methods are based on pairwise relationships~(similarities)~\cite{DBLP:journals/pami/ShiM00,ng2002spectral}. However, the relationships between vertices~(objects) can be more complex than pairwise in real-world applications. In particular, the objects may be grouped together according to different properties or topics. The groups can be viewed as relationships which are typically not pairwise. A good example in industrial domain is the very large scale integration~(\mbox{VLSI}) circuit design~\cite{hagen1992new}. The objects in the circuits are connected in groups via wire nets. Typically, these complex non-pairwise relationships can be represented as hyperedges in hypergraphs~\cite{berge1973graphs}. More specifically, a hypergraph contains a set of vertices and a set of hyperedges, and a hyperedge is an edge that connects at least two vertices in the hypergraph. Note that an ordinary pairwise edge can be treated as a special hyperedge which connects exactly two vertices.

Like NGC, normalized hypergraph cut~(\mbox{NHC})~\cite{catalyurek1999hypergraph,zhou2006learning} tries to divide the vertices into several groups~(clusters) by minimizing the number of hyperedges connecting nodes in different clusters and meanwhile maximizing the number of hyperedges within the clusters. Furthermore, some normalization techniques are also adopted in NHC methods. NHC has been widely used in many applications and attracted more and more attention. For example, NHC has been used to reduce the communication cost and balance the workload in parallel computation such as sparse matrix-vector multiplication~\cite{catalyurek1999hypergraph}. In fact, it is pointed out that hypergraphs can model the communication cost of parallel graph computing more precisely~\cite{catalyurek1999hypergraph}. In addition, the large scale machine learning framework GraphLab~\cite{gonzalez2012powergraph} utilizes similar ideas for distributed graph computing. Due to its wide applications, a lot of algorithms have been proposed for the NHC problem~\cite{bolla1993spectra,karypis1999multilevel,catalyurek1999hypergraph,zhou2006learning,ding2008image,bulo2009game,pu2012hypergraph,anandkumar2014tensor}. These algorithms can be divided into three main categories: heuristic approach, spectral approach, and tensor approach.


The ``hMetis"~\cite{karypis1999multilevel} and ``Parkway"~\cite{trifunovic2004parkway} are two famous tools to solve the NHC problem by using some heuristics. The basic idea is to first construct a sequence of successively coarser hypergraphs. Then the coarsest~(smallest) hypergraph is cut~(partitioned), based on which the cut~(partitioning) of the original hypergraph is obtained by successively projecting and refining the cut results of the former coarser level to the next finer level of the hypergraph.


%
Spectral approach~\cite{bolla1993spectra,zhou2006learning,agarwal2006higher,rodriguez2009laplacian} is also used to solve the NHC problem. In~\cite{bolla1993spectra}, the eigen-decomposition on an unnormalized Laplacian matrix is proposed for hypergraph cut by using ``clique expansion.'' More specifically, each hyperedge is replaced by a clique~(a fully connected subgraph) and the hypergraph is converted to an ordinary graph based on which the Laplacian matrix is constructed. Zhou et al. \shortcite{zhou2006learning} also use ``clique expansion'' as that in~\cite{bolla1993spectra} to convert the hypergraph to an ordinary graph, and then adopt the well-known spectral clustering methods~\cite{ng2002spectral} to perform clustering on the normalized Laplacian matrix. Agarwal et al.~\shortcite{agarwal2006higher} survey various Laplace like operators for hypergraphs and study the relationships between hypergraphs and ordinary graphs. In addition, some theoretical analysis for the hypergraph Laplacian matrix is also provided in~\cite{rodriguez2009laplacian}.

Recent studies~\cite{frieze2008new,cooper2012spectra} suggest using an affinity tensor of order $k$ to represent $k$-uniform hypergraphs, in which each hyperedge  connects to  $k$ nodes exactly. Note that the ordinary graph is a $2$-uniform hypergraph. Tensor decomposition of a high-order normalized Laplacian is then used to solve the NHC problem. Frieze and Kannan \shortcite{frieze2008new} first use a 3-d matrix~(order 3 tensor) to represent a 3-uniform hypergraph and find cliques. Cooper and Dutle \shortcite{cooper2012spectra} generalize such an idea to any uniform hypergraphs and use the high-order Laplacian to partition the hypergraphs.

%

The existing methods mentioned above are limited in some aspects. More specifically,  implementation of the heuristic approaches is simple. However, there does not exist any theoretical guarantee for those approaches. Although proved to be efficient, the spectral approach using ``clique expansion'' does not model the NHC precisely because it approximates the non-pairwise relationships via pairwise ones. Subsequently, some information of the original structure of the hypergraph may be lost due to the expansion. The tensor approach can only be used for uniform hypergraphs, but hypergraphs from real applications are typically not uniform.


In this paper, we propose a novel relaxation approach, which is called \emph{\mbox{relaxed NHC}}~(\mbox{RNHC}), to solve the NHC problem. The main contributions are briefly outlined as follows:
\begin{itemize}
\item We propose a novel model to precisely formulate the \mbox{NHC} problem. Our model can preserve the original structure of general hypergraphs. Moreover, our model does not require the hypergraphs to be uniform.
\item Our formulation for the \mbox{NHC} yields a NP-hard problem.  
 We propose a novel relaxation on the Stiefel manifold. And we efficiently solve the relaxed problem via designing an algorithm based on the Cayley transformation.
\item Experimental results on several large hypergraphs show that \mbox{RNHC} can outperform the state-of-the-art methods.
\end{itemize}

The remainder of this paper is organized as follows.  We first introduce some basic definitions as well as the baseline algorithm in Section~\ref{section:preliminaries}. The RNHC model and our algorithm are proposed in Section~\ref{section:methodology}. In Section~\ref{section:experiment}, we present the experimental results on several VLSI hypergraph benchmarks. Finally, we conclude this paper in Section~\ref{section:conclusion}.

\section{Preliminaries}
\label{section:preliminaries}
In this section, we present the notation and problem definition of this paper. In addition, 
the spectral clustering approach~\cite{zhou2006learning} is also briefly introduced.

\subsection{Notation}
\label{section:notation}

Let $V = \{v_1, v_2, \ldots, v_n\}$ denote the set of $n$ \textit{vertices}~(nodes), and $E= \{e_1, e_2, \ldots, e_m\}$ denote the set of $m$ undirected \textit{hyperedges}. A hyperedge $e \in E$ is a subset of $V$ which might contain more  than two vertices. 
Note that an ordinary edge is still a hyperedge which has exactly two vertices. A \textit{hypergraph} is defined as $H = (V, E)$ with vertex set $V$ and hyperedge set $E$. A hypergraph $H$ can be represented by an $n \times m$ matrix $\B$ with entries $B(v, e) = 1$ if $v \in e$ and $0$ otherwise. We define an $n \times n$ diagonal matrix $\D$ with diagonal entries $D(v, v) = \sum_{e \in E} B(v, e)$ which is the degree of vertex $v$.
Some notations that will be used are listed in Table~\ref{table:notations}.

\begin{table}[t]
\caption{Notation}
\begin{center}
\small
\begin{tabular}{|l|l|}
\hline
Notation	&	Description \\ \hline
$\big|\{\cdot\}\big|$	&	The number of elements in a set \\ \hline
$V = \{v_1, v_2, \ldots, v_n\}$	&	The set of vertices \\ \hline
$|V| = n$	&	The cardinality of $V$ \\ \hline
$E = \{e_1, e_2, \ldots, e_m\}$	&	The set of hyperedges \\ \hline
$|E| = m$	&	The cardinality of $E$ \\ \hline
$D(v,v)$		&	The degree of vertex $v$ \\ \hline
$\1_p$		&	The $p \times 1$ all-one  vector \\ \hline
$\I_p$		&	The $p \times p$ identity matrix \\ \hline
$U_{ij}$    &   The $(i,j)$th element of matrix $\U$ \\ \hline
$\U^T$		&	The transpose of matrix $\U$ \\ \hline
$\U^{-1}$		&	The inverse of matrix $\U$ \\ \hline
$\W = \U \odot \V$	&	$W_{ij} = U_{ij} \times V_{ij}$ \\ \hline
$\W = \U ./ \V$	&	$W_{ij} = U_{ij} / V_{ij}$ \\ \hline
$1_{\{condition\}}$	&	$1$ if the condition is satisfied; \\
					&	$0$ otherwise \\ \hline
$[p]$		&	$\{1,2, \ldots, p\}$ \\ \hline
$\U = \ln(\V)$	&	$U_{ij} = \ln(V_{ij})$ \\ \hline
$\U = \exp(\V)$	&	$U_{ij} = \exp(V_{ij})$ \\ \hline
$||\U||$	&	F-norm of $\U$, i.e., $\sqrt{\sum_{ij}U_{ij}^2}$ \\ \hline
$\tr(\U)$	&	$\sum_{i} U_{ii}$ \\ \hline
$\U = \nabla f(\V)$	&	The gradient of function $f(\V)$, \\
			&	i.e., $U_{ij} = \big( \frac{\partial f(\V)}{\partial V_{ij}} \big)$ \\ \hline
\end{tabular}
\end{center}
\label{table:notations}
\end{table}

\subsection{Normalized Hypergraph Cut}

Let each vertex $v \in V$ be uniquely assigned to a cluster $c \in C$ where $C = \{c_1, c_2, \ldots, c_p\}$ denotes the $p$ clusters. Then each hyperedge spans a set of different clusters. A $p$-way \textit{hypergraph cut} is a sequence of disjoint subsets $c_i \subseteq V$ with $\cup_{i=1}^p c_i = V$. A \emph{min-cut} problem is to find a hypergraph cut such that the number of hyperedges spanning different clusters is minimized.


Now we formally define the normalized hypergraph cut~(\mbox{NHC}) problem. We define the \textit{volume} of cluster $c_i$ to be $vol(c_i)= \sum_{v \in c_i} D(v, v)$. A hyperedge gets one cut if its vertices span two different clusters. Given the set of clusters $C$, the cut of a hyperedge $e$ is defined as 
\[ \sum_{i \in [p]} \sum_{j \in [p], j \neq i} 1_{\{e \cap c_i \neq \emptyset, e \cap c_j \neq \emptyset\}},
\] 
i.e., the number of pairs of different clusters that cut $e$. Then the \emph{cut-value} of a \textit{hypergraph cut} (denoted  $hcut(C)$) is the total number of cuts of all the hyperedges in the hypergraph:
\begin{align*}
hcut(C) = \sum_{e \in E} \sum_{i \in [p]} \sum_{j \in [p], j \neq i} 1_{\{e \cap c_i \neq \emptyset, e \cap c_j \neq \emptyset\}}.
\end{align*}

The cut-value caused by a specific cluster $c_i$ is 
\[ hcut(c_i) = \sum_{e \in E} \sum_{j \in [p], j \neq i} 1_{\{e \cap c_i \neq \emptyset, e \cap c_j \neq \emptyset\}}.
\] 
Then, if we \emph{normalize} the cut-value of each cluster by its volume, we can get the NHC value $nhcut(C)$:
\begin{align}
nhcut(C) = \sum_{i \in [p]} \frac{\sum_{e \in E} \sum_{j \in [p], j \neq i} 1_{\{e \cap c_i \neq \emptyset, e \cap c_j \neq \emptyset\}} }{vol(c_i)}.
\label{equation:nhcut}
\end{align}

The \textit{normalized hypergraph cut}~(NHC) problem is to find a hypergraph cut which can minimize the $nhcut(C)$.
Note that in $nhcut(C)$  the weights of the hyperedges and vertices are assumed to be $1$, 
which means that the hypergraph is unweighted. However, our model, learning algorithm and results of this paper can be easily extended to  weighted hypergraphs, which is not the focus of this paper.

\subsection{Spectral Approach}
\label{section:spectral_approach}

The spectral approach~\cite{zhou2006learning} approximates the NHC via ``clique expansion.''
Each hyperedge $e$ is expanded to a fully connected subgraph with the same edge weight $1/ \big| e \big|$.
Then, the hypergraph is converted to a weighted ordinary graph. The NHC problem is solved by spectral clustering~\cite{ng2002spectral} on the expanded graph. That is, the $p$ smallest eigenvalues will be calculated, and the corresponding eigenvectors will be treated as the new features for the vertices, which will be further clustered via the K-Means algorithm to get the final solution.

In the spectral approach, the number of cuts of a hyperedge is approximated by the edge cut of the corresponding clique, which is the number of edges across different clusters normalized by the total number of vertices in the clique. Note that any pair of vertices in a clique is connected by an edge. Thus, the  hypergraph cut is approximated as
\begin{align*}
approx\_hcut(C) = \sum_{i \in [p]} \sum_{e \in E} \big| e \cap c_i \big| (\big| e \big| - \big| e \cap c_i \big|) / \big| e \big|,
\end{align*}
and the corresponding  NHC is  approximated as
\begin{align*}
approx\_nhcut(C) = \sum_{i \in [p]} \frac{\sum_{e \in E} \big| e \cap c_i \big| (\big| e \big| - \big| e \cap c_i \big|) / \big| e \big|}{vol(c_i)}.
\end{align*}

Although such approximation sounds reasonable, it loses the original structure of the hypergraph and solving the NHC problem becomes indirectly.

\section{Methodology}
\label{section:methodology}

In this section, we provide a novel model to formulate the \mbox{NHC} problem as well as a corresponding relaxed optimization problem. An efficient learning algorithm is then presented to solve it.

\subsection{NHC Formulation}

We first rewrite (\ref{equation:nhcut}) into a matrix form. The solution can be represented as an $n \times p$ matrix $\X$ with entries $X(v, c) = 1$ if $v \in c$ and $0$ otherwise, where $v$ is a vertex and $c$ is a cluster. Note that the columns of $\X$ are mutually orthogonal. Given the assignment of the vertices, the assignment of hyperedges is consequently obtained. A hyperedge occurs in a cluster if and only if at least one of its vertices occurs in the cluster. Thus, the corresponding assignment of hyperedges can be represented as a $p \times m$ matrix $\S$ with entries $S(c, e) = 1$ if $\exists v \in c$ such that $e \ni v$ and $S(c, e) = 0$ otherwise, where $e$ is a hyperedge. Note that $\S = sgn(\X^T\B)$ where $sgn()$ is the element-wise sign function.
The $i$th row of $\S = [\s_1, \s_2, \ldots, \s_p]^T$  represents the hyperedges belonging to the corresponding cluster $c_i$. Note that $\s_i^T\s_j$ represents the number of hyperedges between the two clusters $c_i$ and $c_j$,  namely, the cut associated with the two clusters. And the $i$the diagonal element of the matrix $\X^T\D\X$ corresponds to the volume of the $i$th cluster.

Thus, the NHC problem can be represented as:
\begin{align}
& \min\limits_{\X} \tr(\S\S^T(\1_p\1_p^T-\I_p)(\X^T\D\X)^{-1}).
\label{equation:nhcut_matrix_origin}
\end{align}
Recall that  $\X$ is column orthogonal. 
To express such constraints, we normalize the columns of $\X$ to get $\bar\X$ such that $\bar{X}(v, c) = X(v, c) / \sqrt{\sum_{i=1}^n X(i, c)}$. 
Thus, we have $\bar\X^T\bar\X = \I_p$.
We also let $\bar\S = sgn(\bar\X^T\B)$.
To simplify the representation, we assume the degrees of the vertices are roughly the same.
Then, we can rewrite the optimization problem in~\eqref{equation:nhcut_matrix_origin} as follows:
\begin{align}
& \min\limits_{\bar\X} \tr(\bar\S^T(\1_p\1_p^T-\I_p)\bar\S), \mbox{ s.t. } \bar\X^T\bar\X = \I_p. \label{equation:nhcut_matrix_simple1}
\end{align}

\subsection{Relaxation}

The problem in~(\ref{equation:nhcut_matrix_simple1}) is NP-hard, which is intractable to solve because of the quadratic form in the objective function and the occurrence of the sign function. To make the problem tractable, we will simplify the expression and use elementary matrix operations to obtain $\S$ and the corresponding $\bar\S$ before we relax the problem.

First, we simplify the objective function. Note that the quadratic form in (\ref{equation:nhcut_matrix_simple1}) can be rewritten in a summation form
\begin{align}
\sum_{e \in E} p_e \times (p_e - 1),
\label{equation:objfun_origin}
\end{align}
 where $p_e$ is the number of clusters that hyperedge $e$ occurs, or $p_e = \sum_{c \in C} S(c, e)$. Obviously, minimizing~(\ref{equation:objfun_origin}) is equivalent to minimize
 \begin{align}
\sum_{e \in E} p_e,
\label{equation:objfun_simple}
\end{align}
which leads to a simpler expression of the problem:
\begin{align}
& \min\limits_{\bar\X} \sum_{ij}\bar S_{ij}, \mbox{ s.t. } \bar\X^T\bar\X = \I_p. \label{equation:nhcut_matrix_simple2}
\end{align}

Next, we simplify the expression of $\S$. Recall the definition of $\S$, that is, $S(c, e) = 1$ if and only if $\exists v \in e$ such that $v \in c$. In that case, each element of $\S$ can be expressed as $S(c, e) = \max_{v \in e} X(v, c)$. And the minimizer does not change if we substitute $\X$ with $\bar{\X}$.

Now we further relax the problem. We simply relax $\bar{\X}$ to a $n \times p$ real matrix satisfying $\bar\X^T\bar\X = \I_p$. To make the objective function differentiable, we replace the maximum function by the log-sum-exp function, which is a differentiable approximation of the maximum function. We denote the relaxed $\bar{\S}$ by $\hat{\S}$. Then, we have $\hat{S}(c, e) = \frac{1}{\alpha} \ln\big\{ \sum_{v \in e} \exp[\alpha \bar{X}(v, c)] \big\}$, where $\alpha$ is an enough large  parameter. When $\alpha$ gets larger, the approximation gets closer to the maximum function. Moreover,  $\bar{\S}$ can be written in a matrix form:
\begin{align}
\hat{\S} = \frac{1}{\alpha} \ln \big[ \exp(\alpha\bar{\X})^T\B \big].
\label{Shat}
\end{align}
Then, we can relax the NHC problem as follows:
\begin{align}
& \min\limits_{\bar\X} \sum_{ij}\hat S_{ij}, \mbox{ s.t. } \bar\X^T\bar\X = \I_p. \label{equation:nhcut_matrix_simple3}
\end{align}

\subsection{Learning Algorithm}

Now we devise an learning algorithm to solve the relaxed optimization problem in~(\ref{equation:nhcut_matrix_simple3}). 
Since the objective function is minimized under the orthogonality constraint,
the corresponding feasible set $\mathcal{M}_n^p = \big\{ \bar\X \in \mathbb{R}^{n \times p}\big| \bar\X^T\bar\X = \I_p  \big\}$ is called the Stiefel manifold. There are  algorithms in~\cite{edelman1998geometry,wen2013feasible}, which have been proposed to deal with such kinds of constraints. Note that the optimization problem with orthogonality constraints is non-convex, which means there is no guarantee to obtain a global minimizer.  To find a local minimizer on the Stiefel manifold, we introduce the Cayley transformation~\cite{wen2013feasible} to devise the learning  algorithm.

Given any feasible $\bar\X$ and the differentiable objective function $f(\bar\X) = \sum_{ij}\hat S_{ij}$, where $\hat\S$ is defined in (\ref{Shat}), we compute the gradient matrix with respect to $\bar\X$:
\begin{align}
\G = \big\{ \B\big[(\1_p\1_m^T) ./ (\widetilde{\X}^T\B)\big]^T \big\} \odot \widetilde{\X},
\label{equation:gradient}
\end{align}
where $\widetilde{\X} = \exp(\alpha\bar\X)$.

Denote 
\begin{align}
\A = \G\X^T - \X\G^T,
\label{equation:A}
\end{align}
which is skew-symmetric. 
Then we have the Cayley transformation
\begin{align}
\Q = (\I+\frac{\tau}{2}\A)^{-1}(\I+\frac{\tau}{2}\A).
\label{equation:Q}
\end{align}
And the new trial point starting from $\bar\X$ will be searched on the curve
\begin{align}
\Y(\tau) = \Q\bar\X.
\label{equation:Ytau}
\end{align}
Note that $\Y(0) = \bar\X$, $\Y(\tau)^T\Y(\tau) = \bar\X^T\bar\X$ for all $\tau \in \mathbb{R}$, and $\Y(\tau)$ is smooth in $\tau$. Furthermore, $\{\Y(\tau)\}_{\tau \geq 0}$ is a descent path because $\frac{\mathrm{d}}{\mathrm{d}\tau} \Y(0)$ equals the projection of $(-\G)$ into the tangent space of $\mathcal{M}_n^p$ at $\bar\X$.

With all the properties above, we can solve the relaxed problem by a gradient descent algorithm on the curve and discretize the solution via the K-Means clustering algorithm. We present the whole learning procedure in Algorithm~\ref{algorithm:nhcut}.

\begin{algorithm}[htbp!]
\caption{A relaxed learning algorithm for the NHC problem~(RNHC)}
\begin{algorithmic}[1]
\REQUIRE
A hypergraph $H$ and the corresponding matrix $\B$;
number of clusters $p$;
maximal number of iterations $T$;
a stopping criterion $\epsilon$;
the parameter $\alpha$ of log-sum-exp function

\ENSURE
A binary matrix $\X$.

\STATE Initialization: pick an arbitrary orthogonal matrix $\bar{\X}_0 \in \mathbb{R}^{n \times p}$, set $t \leftarrow 0$.
\FOR{$t \leftarrow \{1, \cdots, T\}$}
	\STATE Generate $f(\bar\X_t)$, $\G$, and $\A$ according to (\ref{equation:nhcut_matrix_simple3}), (\ref{equation:gradient}), (\ref{equation:A}).
	\STATE Compute the step size $\tau_t$ by using line search along the path $\Y(\tau)$ defined by (\ref{equation:Ytau}).
	\STATE Set $\bar\X_{t+1} \leftarrow \Y(\tau_t)$.
	\IF{$\parallel \nabla f(\bar\X_{t+1}) \parallel \leq \epsilon$}
		\STATE Stop.
	\ENDIF
\ENDFOR
\STATE $\X = \textit{K-Means}(\bar{\X}_t, p)$.
\end{algorithmic}
\label{algorithm:nhcut}
\end{algorithm}

\subsection{Complexity Analysis}

We now study the time complexity and storage cost of our algorithm.

\subsubsection{Time Complexity}

We analyze the time complexity step by step. The  flops for computing the objective function, the gradient matrix $\G$ and the corresponding skew-symmetric matrix $\A$ are $O(mnp)$, $O(mnp)$ and $O(n^2p)$, respectively. The computation of $\Q$ in (\ref{equation:Q}) needs to compute the inverse of $(\I+\frac{\tau}{2}\A)$. According to the Sherman-Morrison-Woodbury formula, when $p$ is much smaller than $n/2$, we only need to compute the inverse of a $2p \times 2p$ matrix. Thus, the computation of $\Y(\tau)$ needs $8np^2 +O(p^3)$. For a different $\tau$, updating $\Y(\tau)$ needs $4np^2 +O(p^3)$. Note that we always have $m \geq n$.

Putting all the above components together, assuming the number of iterations to be $T$ and ignoring the K-Means step, the time complexity of our algorithm is $O(T \times mnp)$.

Ignoring the K-Means step, the time complexity of the spectral approach~\cite{zhou2006learning} is $O(mnp)$ for computing the $p$ largest singular values and singular vectors.

Although the time complexity of our algorithm seems larger than the spectral approach~\cite{zhou2006learning}, the number of iterations $T$ can be tuned by changing the stopping criterion. Note that $T$ is usually small enough, so it can be viewed as a constant, which yields the same time complexity $O(mnp)$ as the spectral approach. Moreover, a smaller $T$ may lead to a constant factor of reduction for the time complexity.

\subsubsection{Storage Cost}

Ignoring the K-Means step, the largest matrix constructed in our algorithm is $\hat{\S}$. So the storage cost is $O(mp)$.

\section{Experiment}
\label{section:experiment}
In this section, empirical evaluation on real hypergraphs is used to verify the effectiveness of our algorithm. Our experiment is taken on a workstation with Intel E5-2650-v2 2.6GHz~($2 \times 8$ cores) and 128GB of DDR3 RAM.

\subsection{Datasets and Baselines}
The hypergraphs used in our experiment are from a set of hypergraph benchmarks for clustering and partitioning in VLSI domain from the ISPD98 Circuit Benchmark Suite~\footnote{\url{http://vlsicad.ucsd.edu/UCLAWeb/cheese/ispd98.html}}. There are totally $18$ hypergraphs in the dataset. The $12$ largest ones of them are chosen for our experiment. The information of the datasets is shown in Table~\ref{table:hypergraphs}.
\begin{table}[h!]
\caption{Hypergraph datasets}\label{table:hypergraphs}
\begin{center}
\begin{tabular}{|r|r|r|}
\hline
Hypergraph & \# Vertices & \# Hyperedges \\ \hline
ibm07 & 45926 & 48117 \\ \hline
ibm08 & 51309 & 50513 \\ \hline
ibm09 & 53395 & 60902 \\ \hline
ibm10 & 69429 & 75196 \\ \hline
ibm11 & 70558 & 81454 \\ \hline
ibm12 & 71076 & 77240 \\ \hline
ibm13 & 84199 & 99666 \\ \hline
ibm14 & 147605 & 152772 \\ \hline
ibm15 & 161570 & 186608 \\ \hline
ibm16 & 183484 & 190048 \\ \hline
ibm17 & 185495 & 189581 \\ \hline
ibm18 & 210613 & 201920 \\ \hline
\end{tabular}
\end{center}
\end{table}

In our experiment, we adopt the spectral approach~\cite{zhou2006learning} mentioned in Section~\ref{section:spectral_approach} as the baseline. Our method in Algorithm~\ref{algorithm:nhcut} is named as relaxed normalized hypergraph cut~(RNHC). In all the experiments, we pick parameter $\alpha = 100$, the maximal number of iterations $T = 1000$, and the stopping criterion $\epsilon = 10^{-9}$ for our algorithm. The reason why we do not compare our algorithm with the heuristic approaches such as hMetis and Parkway is that we only focus on solving NHC problem via optimization in this paper. And the spectral approach is the algorithm most related to our work.

\subsection{Clustering Visualization}
\begin{figure}[!htb]
\centering
\subfigure[Spectral approach]{\label{fig:visualization_a}\includegraphics[width=0.23\textwidth,height=0.22\textwidth]{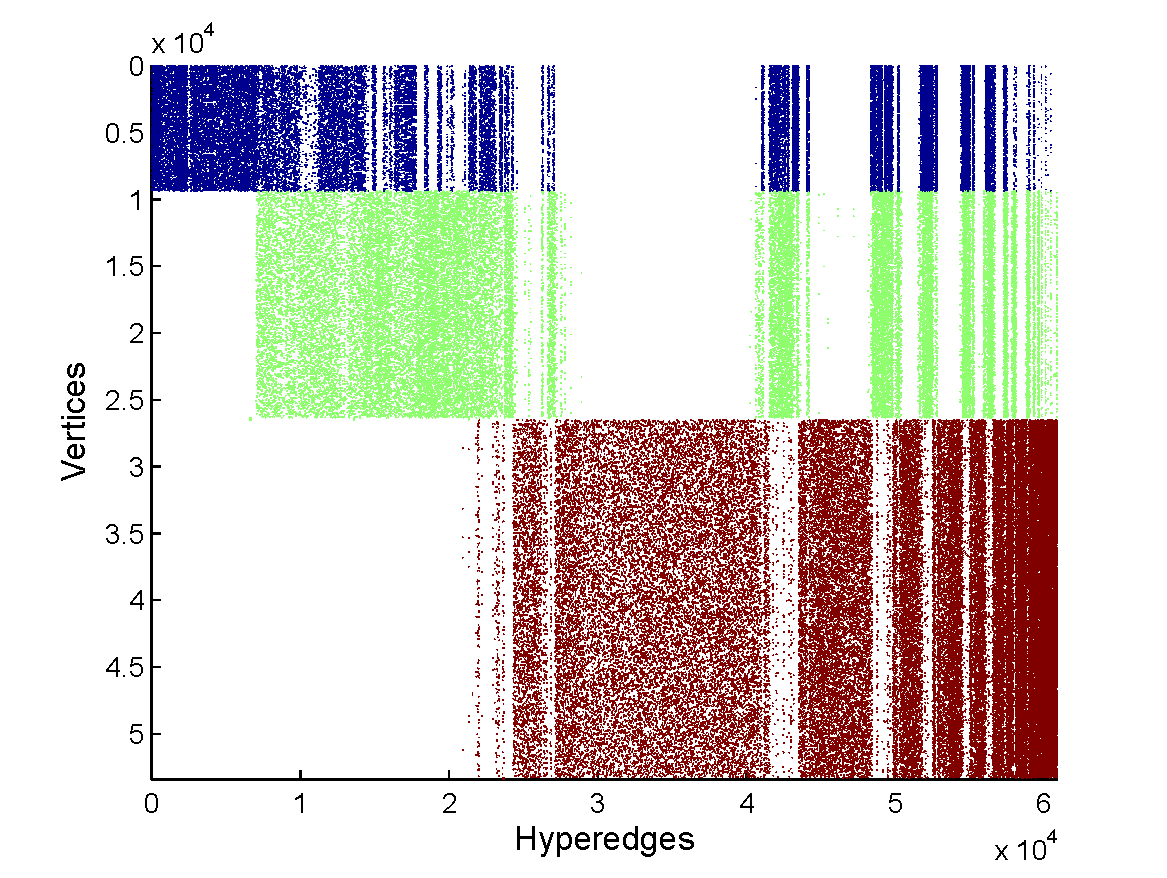}}
\subfigure[RNHC]{\label{fig:visualization_b}\includegraphics[width=0.23\textwidth,height=0.22\textwidth]{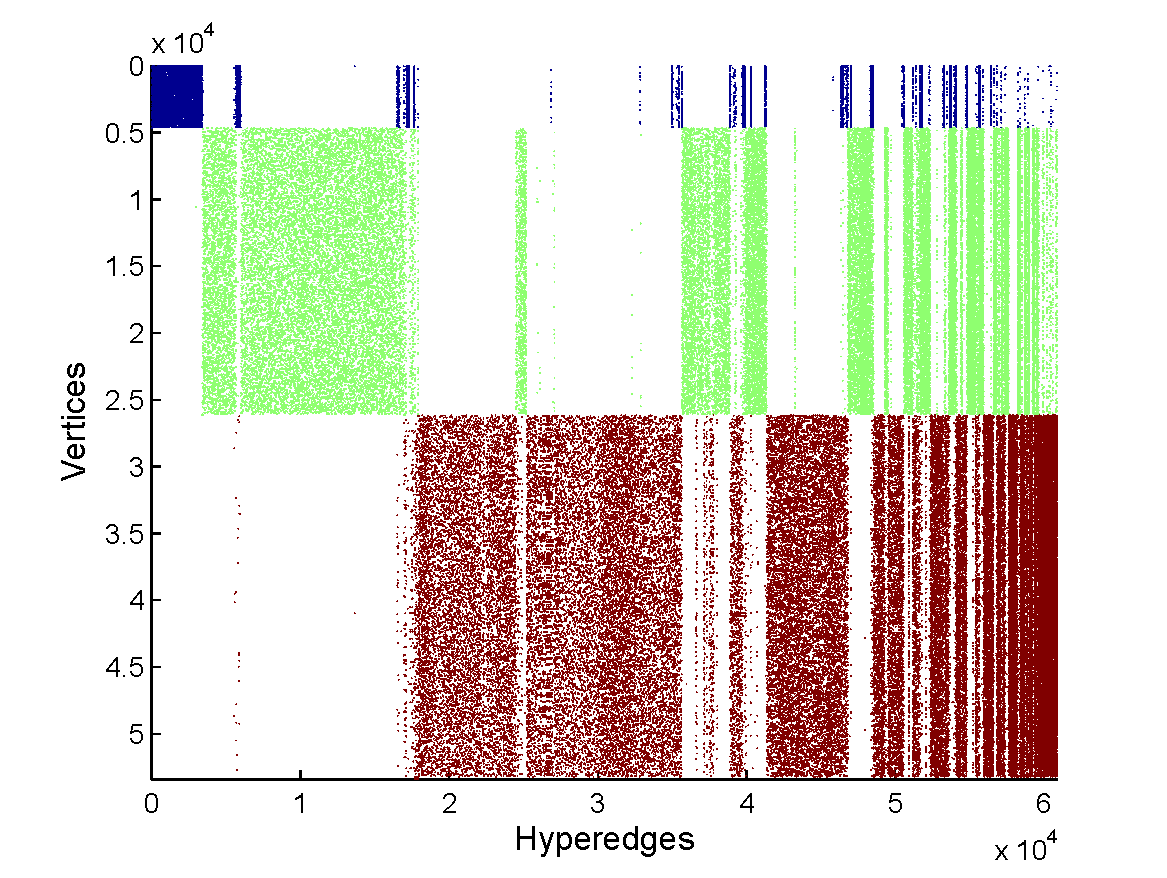}}
\caption{Clustering visualization on matrix $\B$ of ibm09. There are 3 
 clusters indicated by blue, green and red. Note that each column in the 
 image represents a hyperedge. The number of colors of a column indicates 
 the number of clusters that the corresponding hyperedge spans. It can be 
 seen from Figure~\ref{fig:visualization_a} that the blue and green 
 clusters share many hyperedges~(columns), which implies a bad 
 clustering. And in Figure~\ref{fig:visualization_b}, less overlapping 
 columns are produced by RNHC, which means a better clustering. }
\label{fig:visualization}
\end{figure}

We visualize the clustering produced by the spectral approach and RNHC on ibm07 in Figure~\ref{fig:visualization}. In the figure, we illustrate the matrix $\B$ defined in Section~\ref{section:notation}. Each row of $\B$ represents a vertex and  each column represents a 
 hyperedge. A non-blank point located at $(x, y)$ in the figure implies that the $y$th vertex belongs to the $x$th hyperedge. The number of clusters is 3. Different colors indicate different clusters. The vertices are rearranged such that vertices in the same cluster will be grouped together. And the hyperedges in both Figures~\ref{fig:visualization_a} and \ref{fig:visualization_b} are arranged in the same order. In a better clustering, there should be less overlapping columns~(hyperedges) between clusters.

\subsection{Accuracy Comparison}
We evaluate the number of clusters from $2$ to $8$. For each trial, both algorithms will be tested for 40 times and the best NHC value will be picked for comparison. The reason why we compare the best NHC value instead of the average performance is that both algorithms utilize the K-Means algorithm for final clustering, whose result depends on the starting point. Sometimes K-Means simply fails to obtain $p$ clusters, which means that some of the clusters are empty. Moreover, K-Means occasionally produces extremely bad NHC values because of some bad starting point. Such failures may make the average performance meaningless. Furthermore, it is also hard to tell which trial fails and which one succeeds.

\begin{figure*}[!tb]
\centering
\subfigure[ibm07]{\includegraphics[width=0.22\textwidth,height=0.2\textwidth]{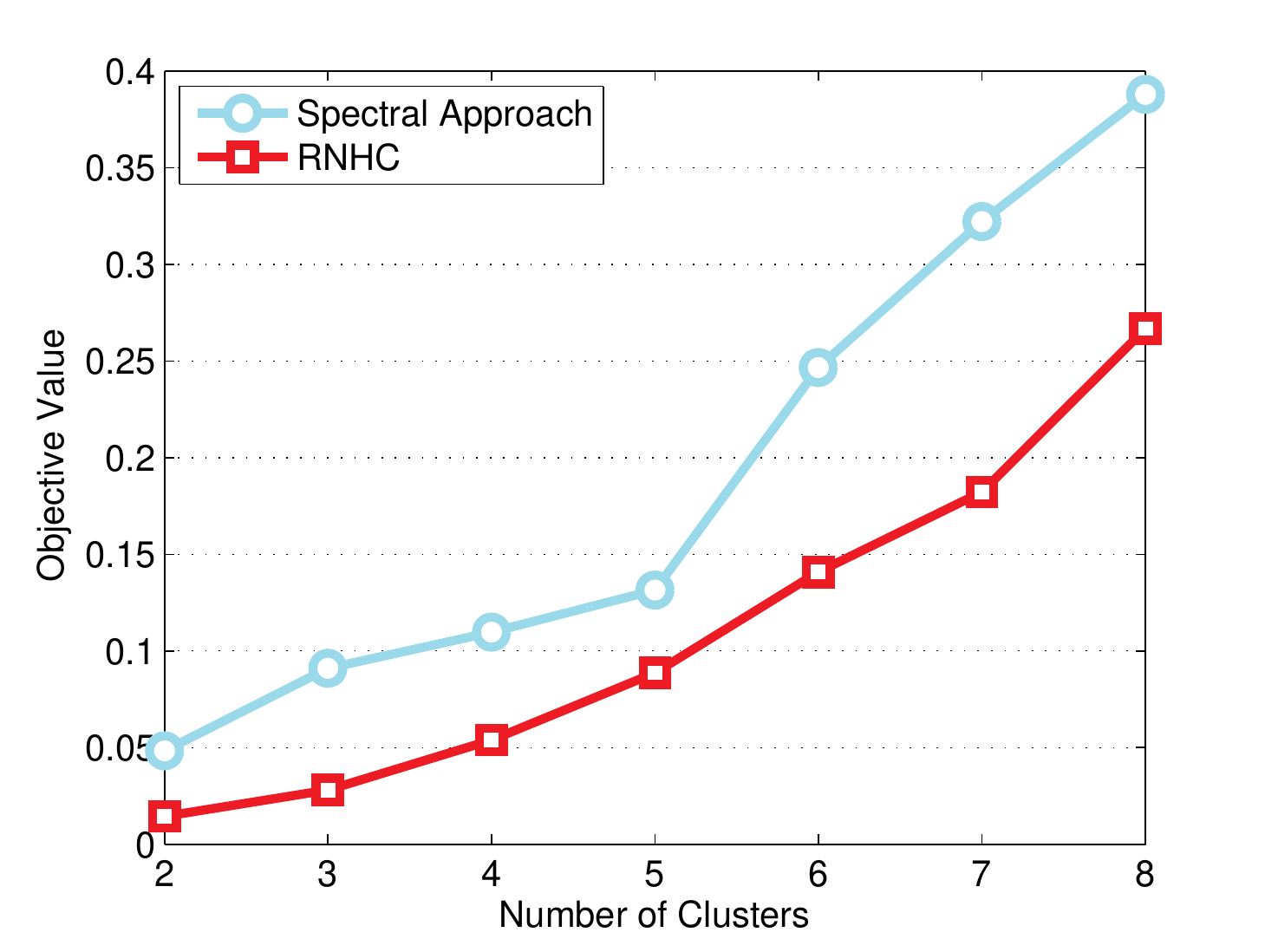}}
\subfigure[ibm08]{\includegraphics[width=0.22\textwidth,height=0.2\textwidth]{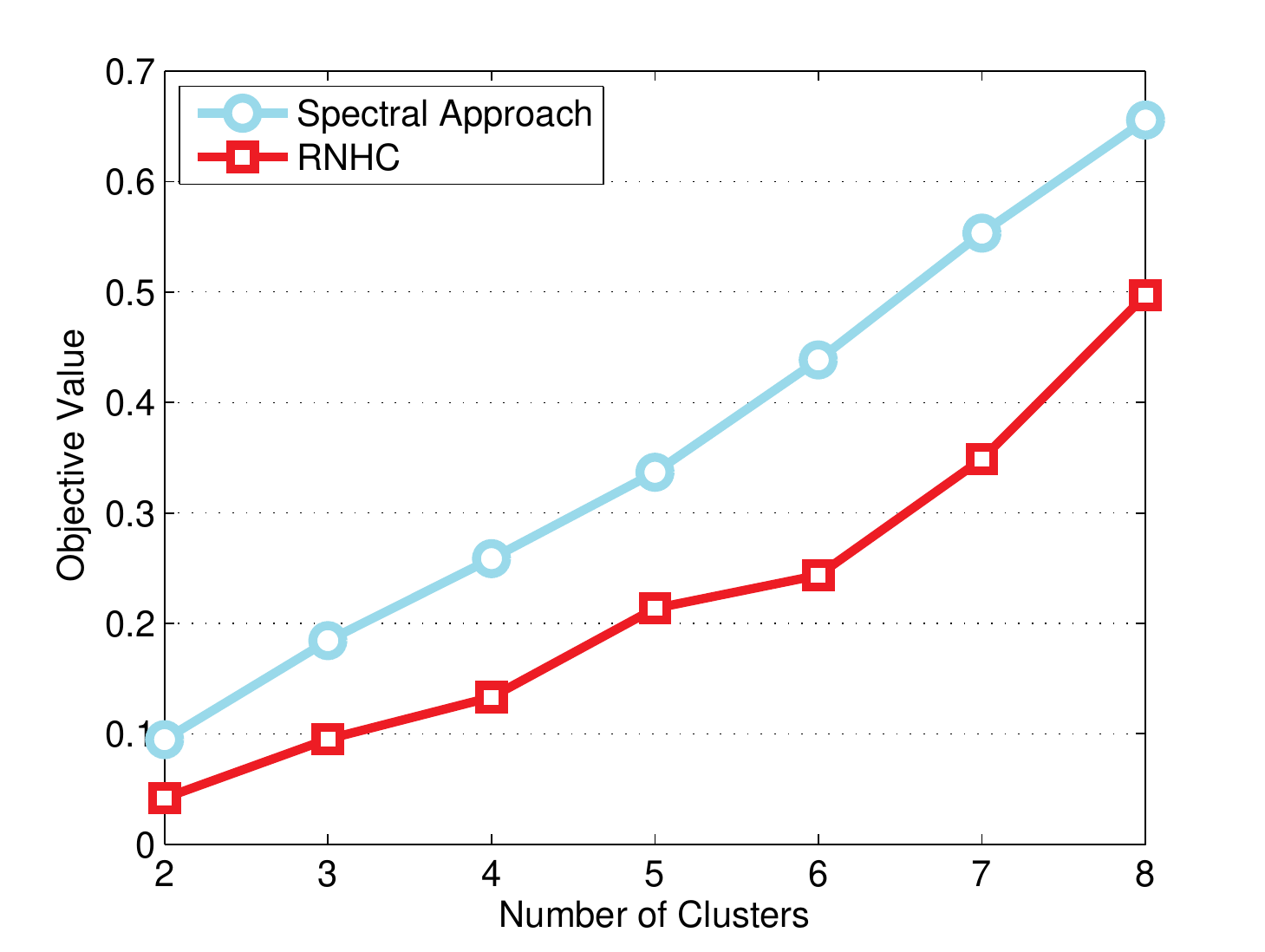}}
\subfigure[ibm09]{\includegraphics[width=0.22\textwidth,height=0.2\textwidth]{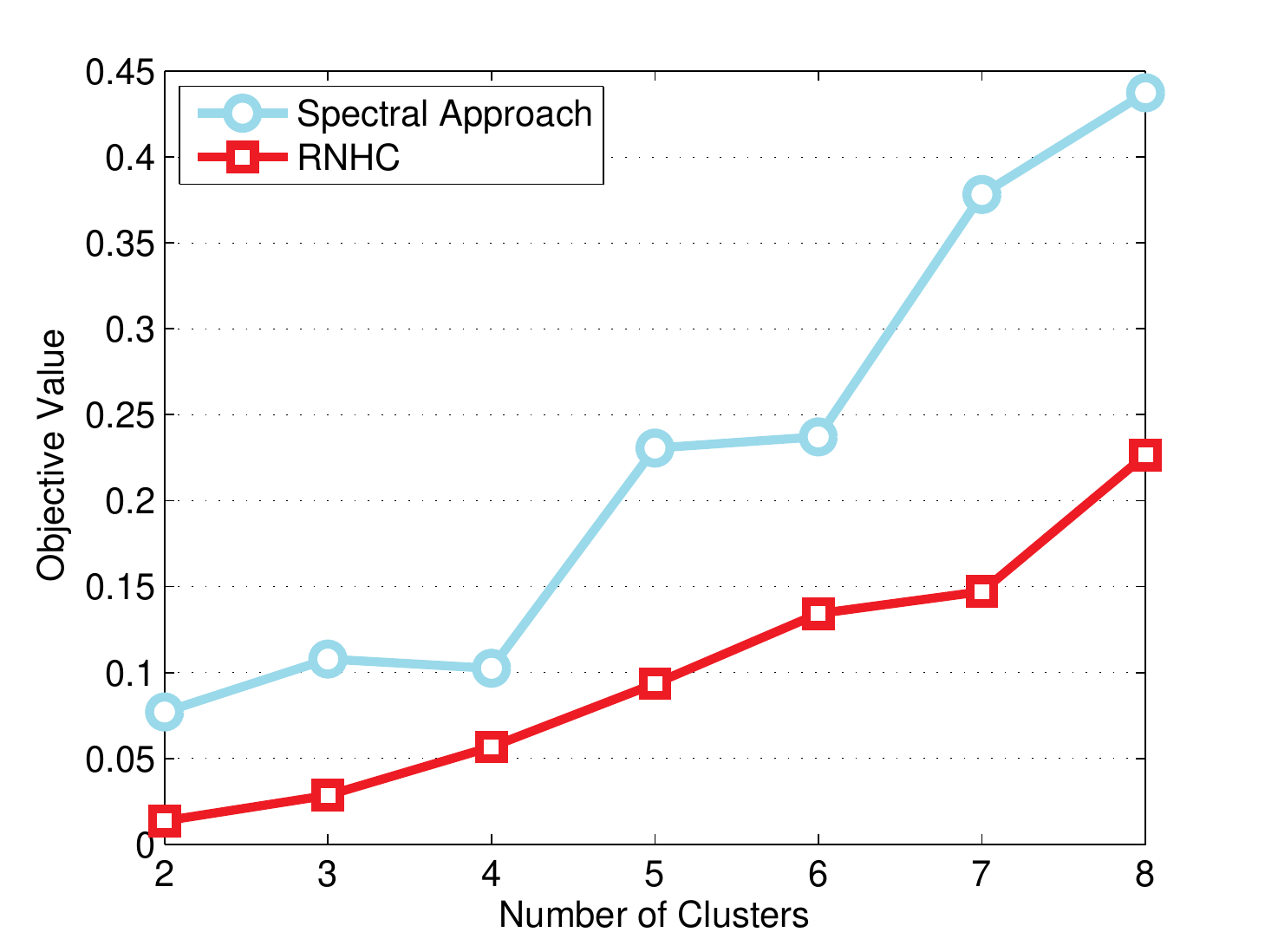}}
\subfigure[ibm10]{\includegraphics[width=0.22\textwidth,height=0.2\textwidth]{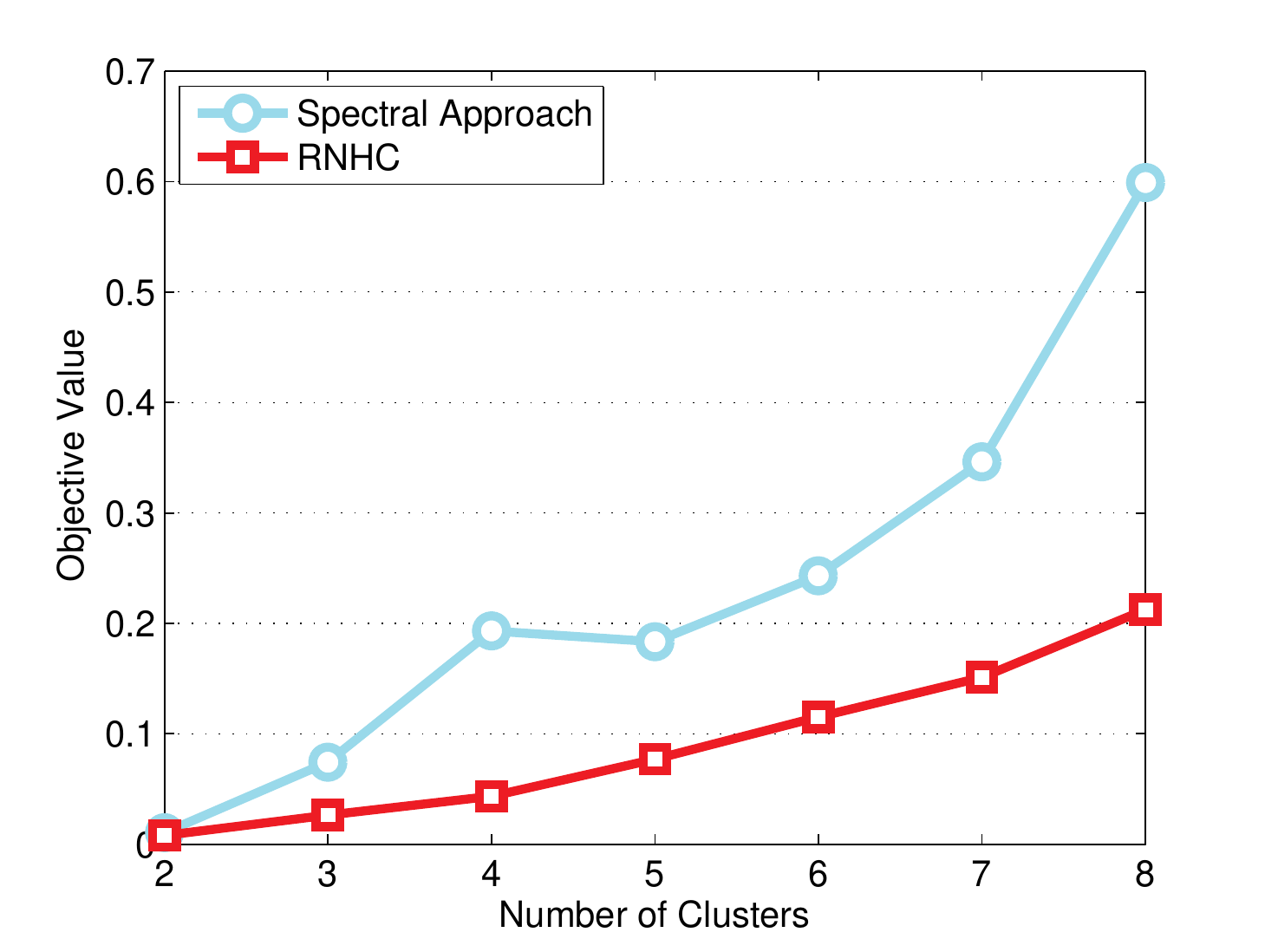}}
\subfigure[ibm11]{\includegraphics[width=0.22\textwidth,height=0.2\textwidth]{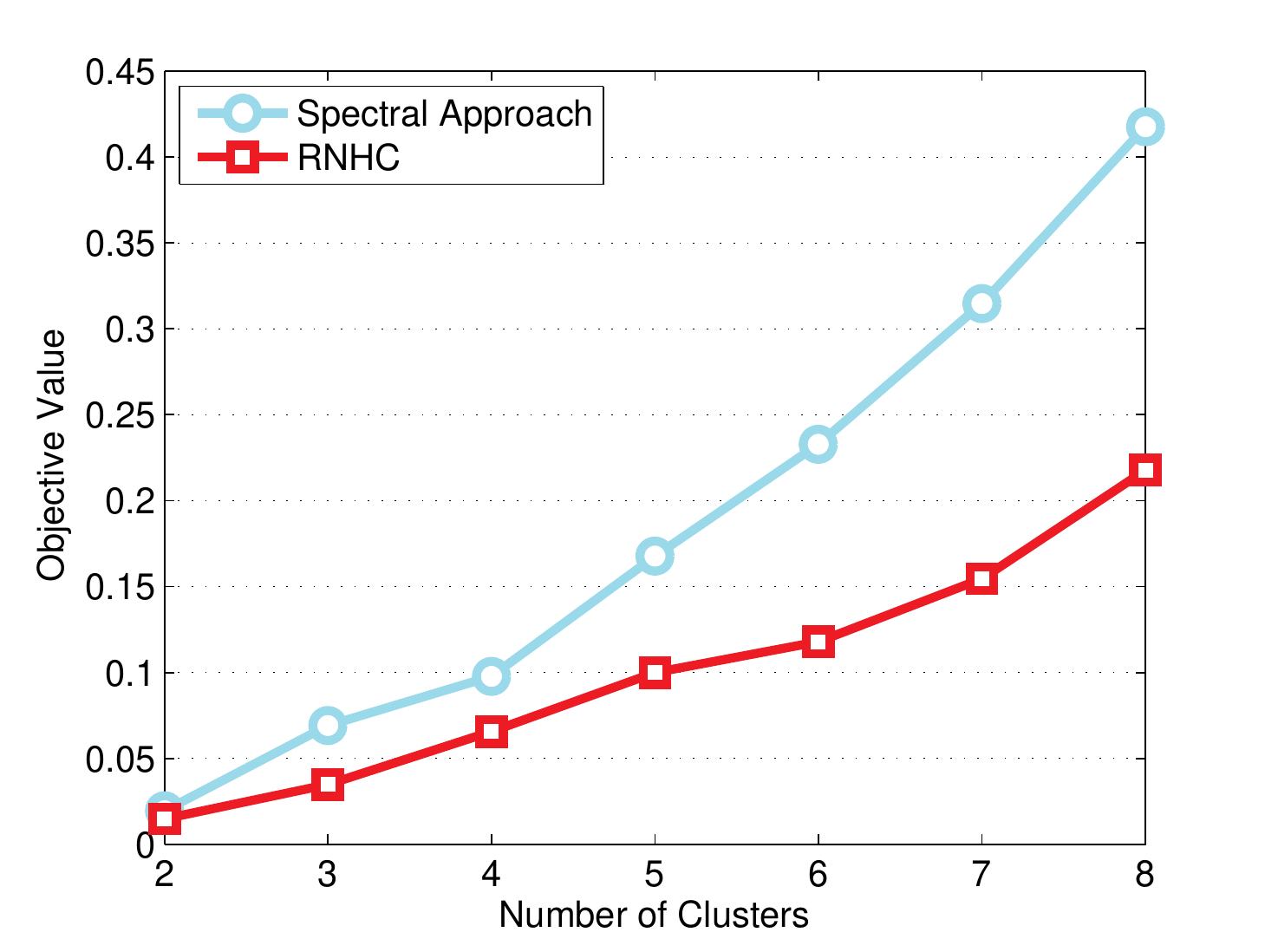}}
\subfigure[ibm12]{\includegraphics[width=0.22\textwidth,height=0.2\textwidth]{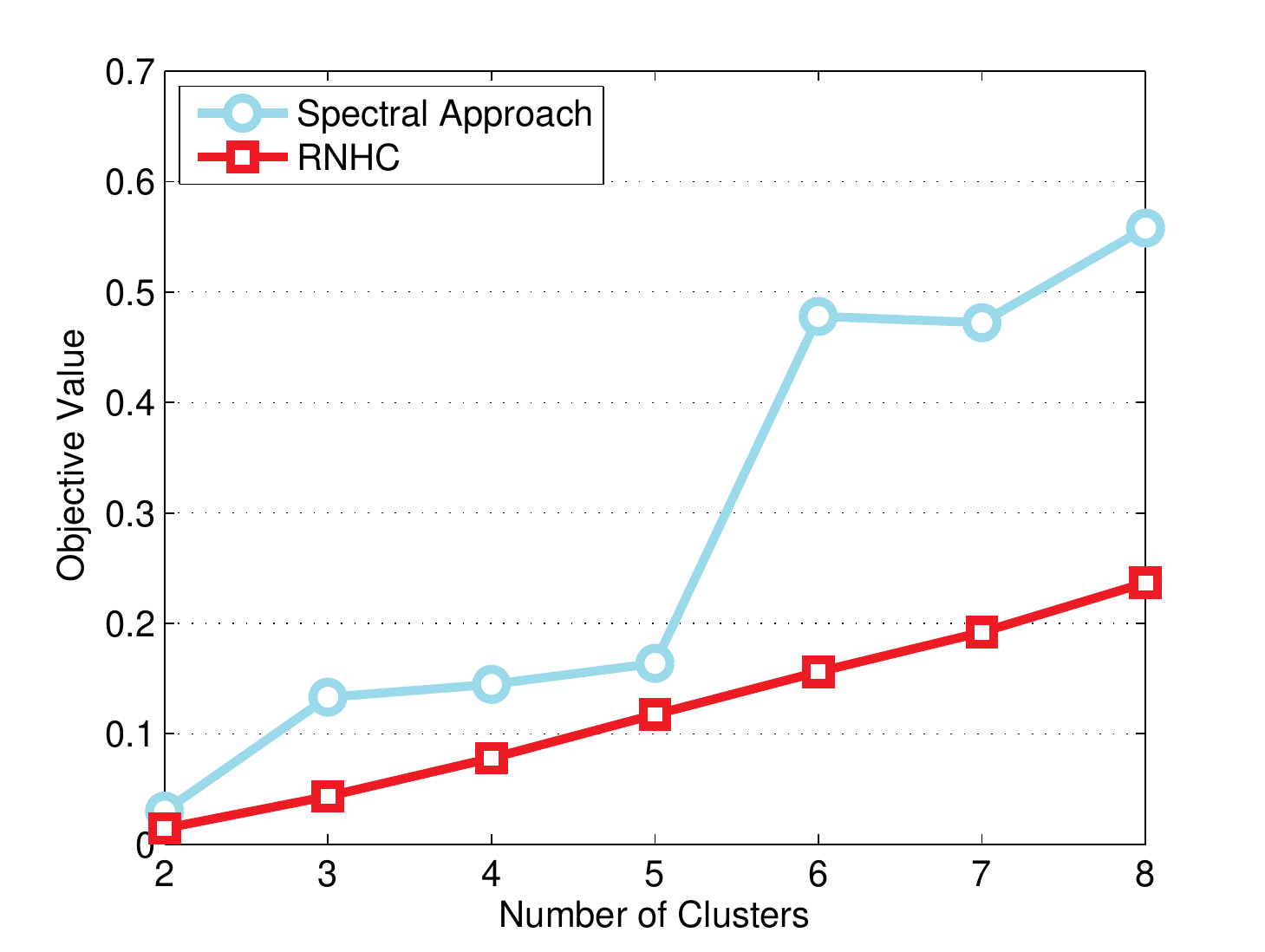}}
\subfigure[ibm13]{\includegraphics[width=0.22\textwidth,height=0.2\textwidth]{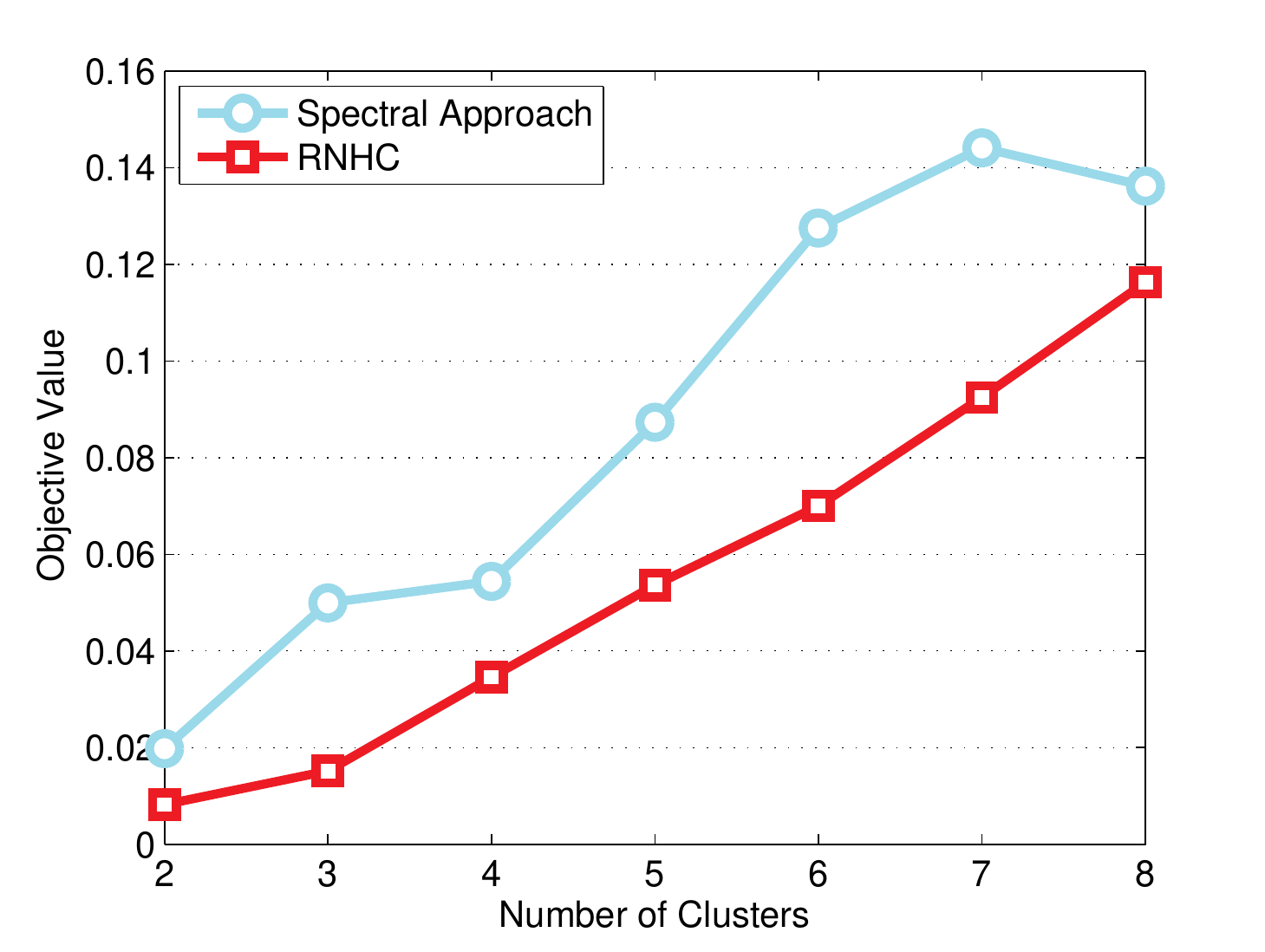}}
\subfigure[ibm14]{\includegraphics[width=0.22\textwidth,height=0.2\textwidth]{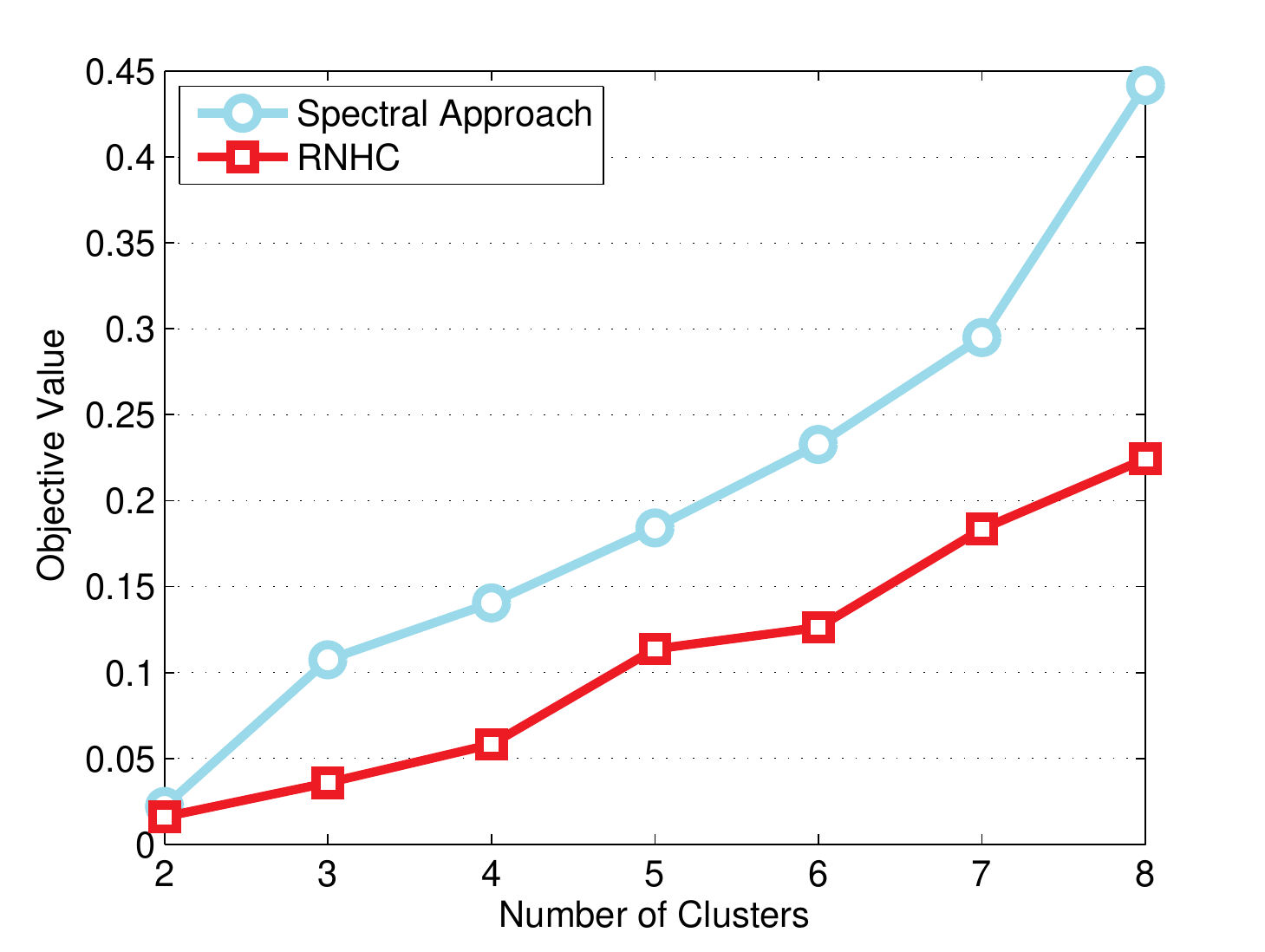}}
\subfigure[ibm15]{\includegraphics[width=0.22\textwidth,height=0.2\textwidth]{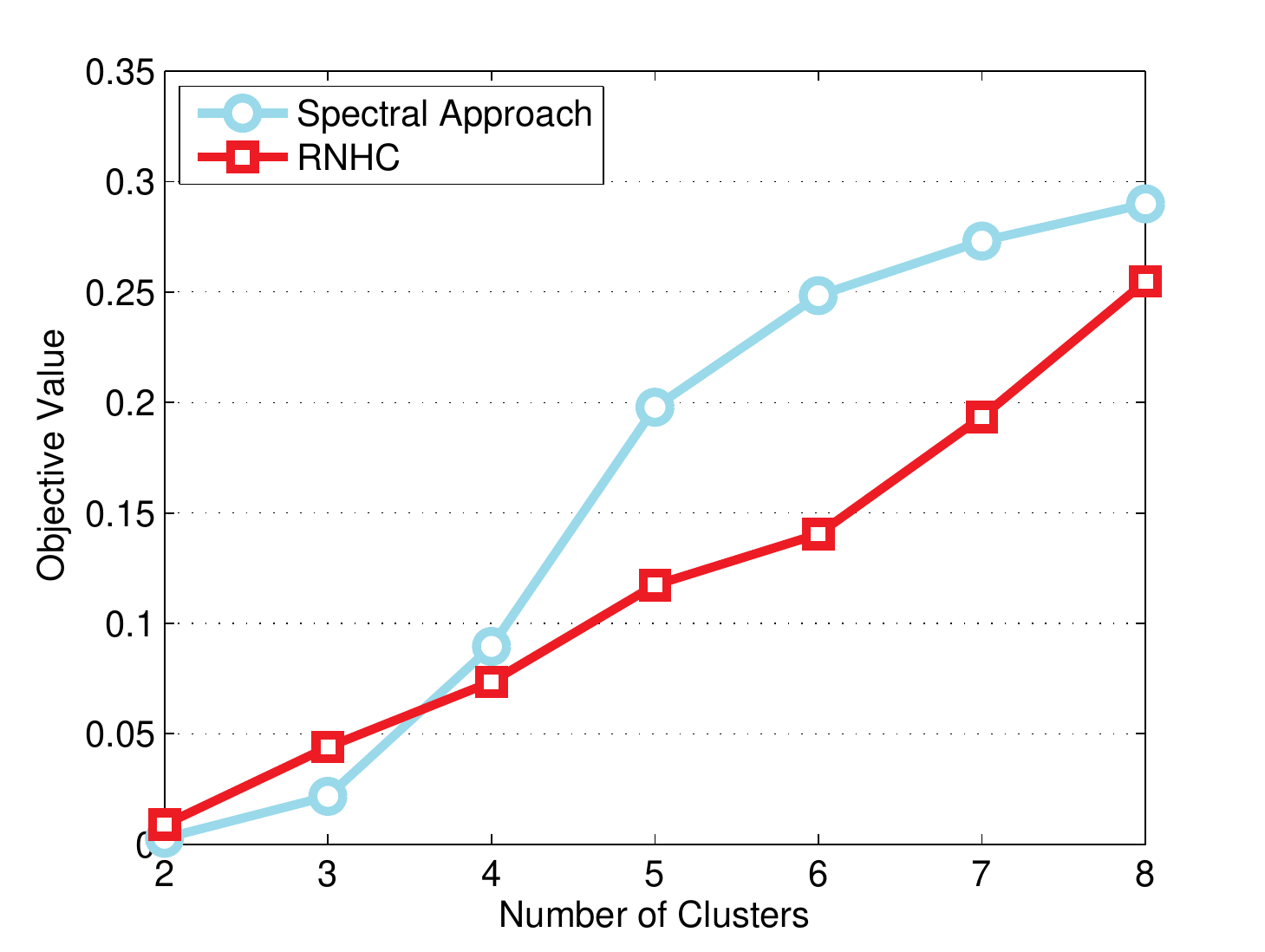}}
\subfigure[ibm16]{\includegraphics[width=0.22\textwidth,height=0.2\textwidth]{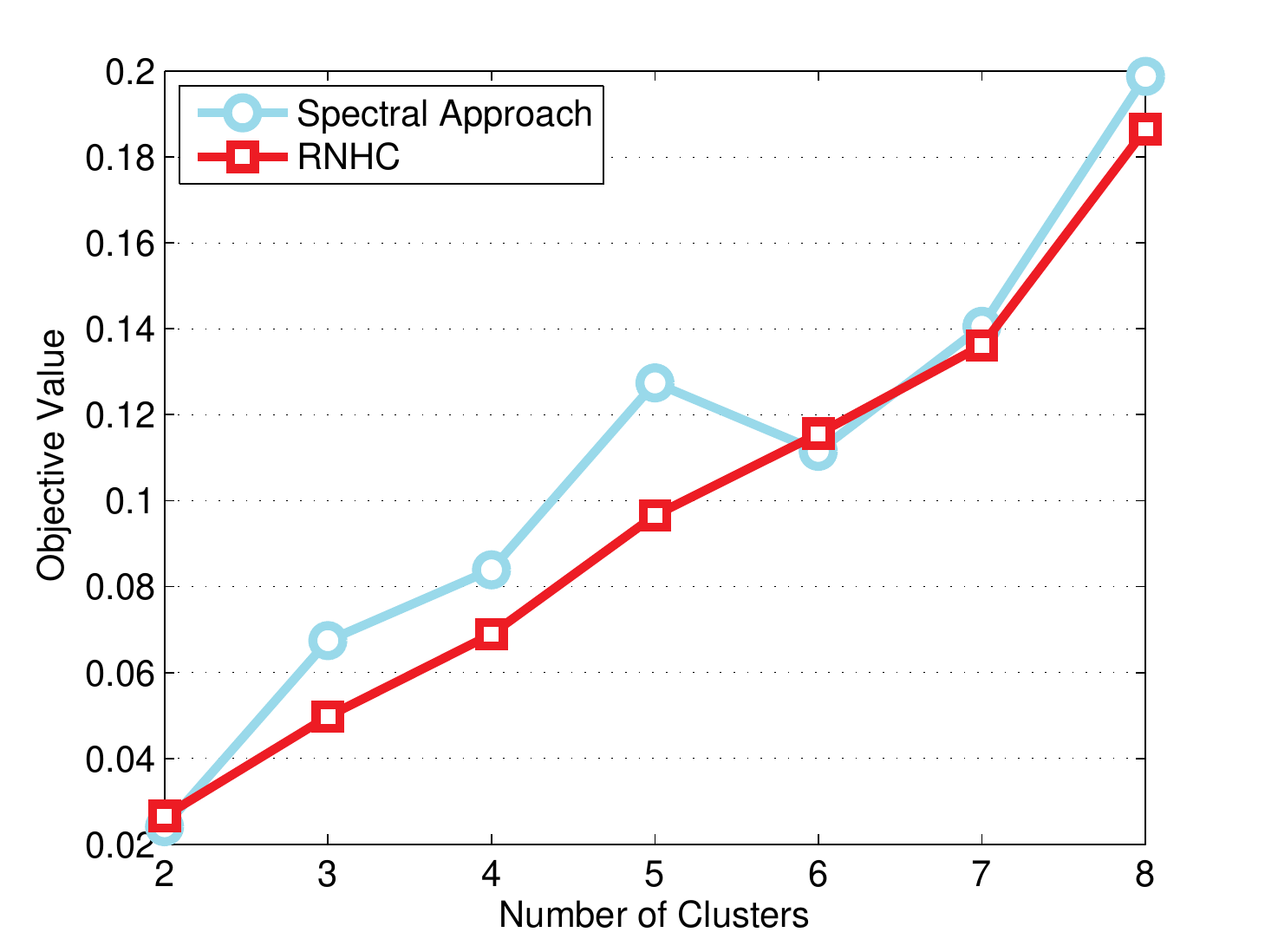}}
\subfigure[ibm17]{\includegraphics[width=0.22\textwidth,height=0.2\textwidth]{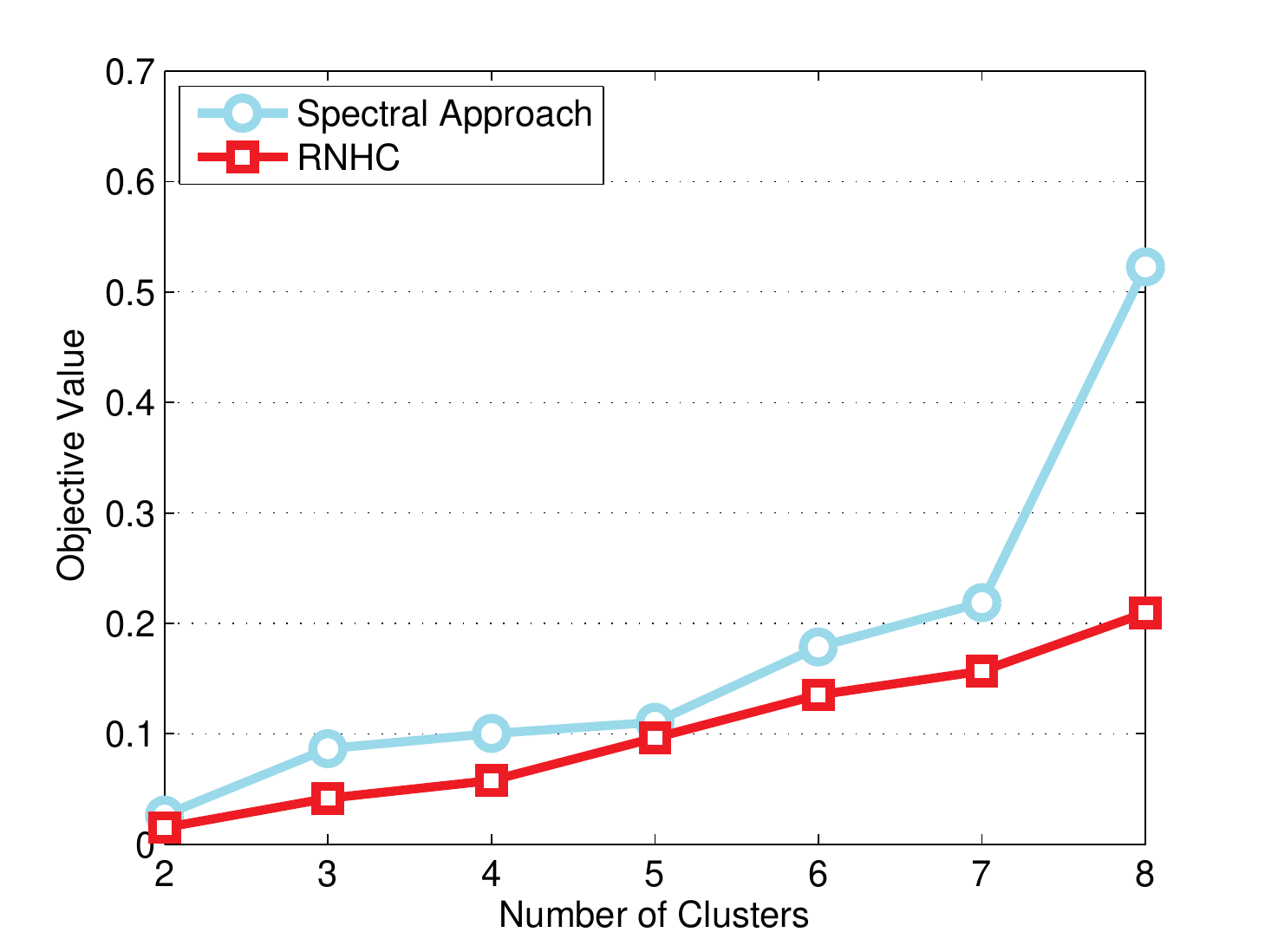}}
\subfigure[ibm18]{\includegraphics[width=0.22\textwidth,height=0.2\textwidth]{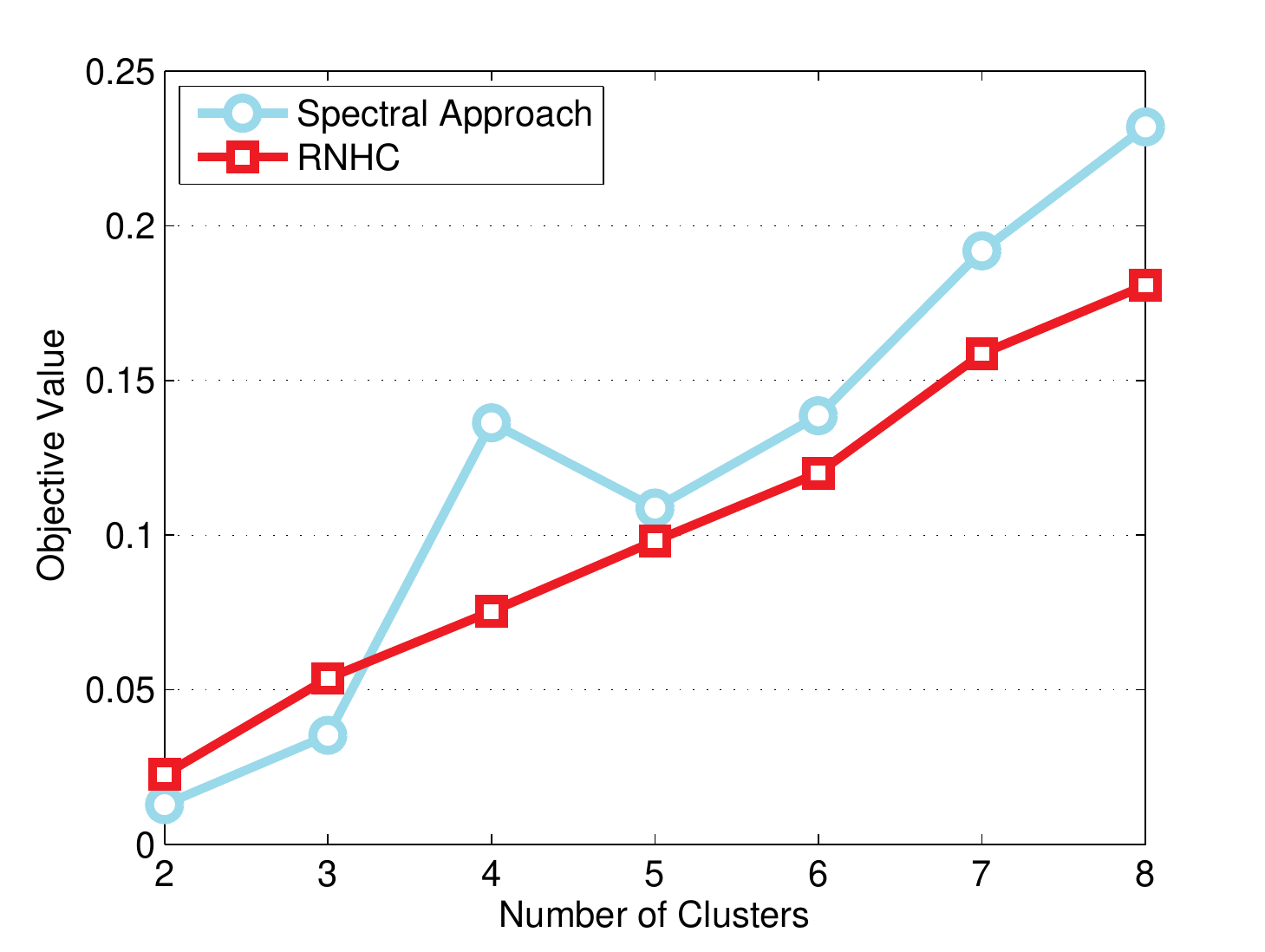}}
\caption{Objective value of clustering on ibm07-18}
\label{fig:obj1}
\end{figure*}

We test the objective value of NHC in~(\ref{equation:nhcut_matrix_origin}) for each algorithm. The comparison of the objective value is shown in Figure~\ref{fig:obj1}. Note that a smaller objective value implies a better NHC. It can be seen that our algorithm produces a better objective value in most cases.

\subsection{Speed Comparison}

\begin{figure*}[!htb]
\centering
\subfigure[ibm15]{\includegraphics[width=0.22\textwidth,height=0.20\textwidth]{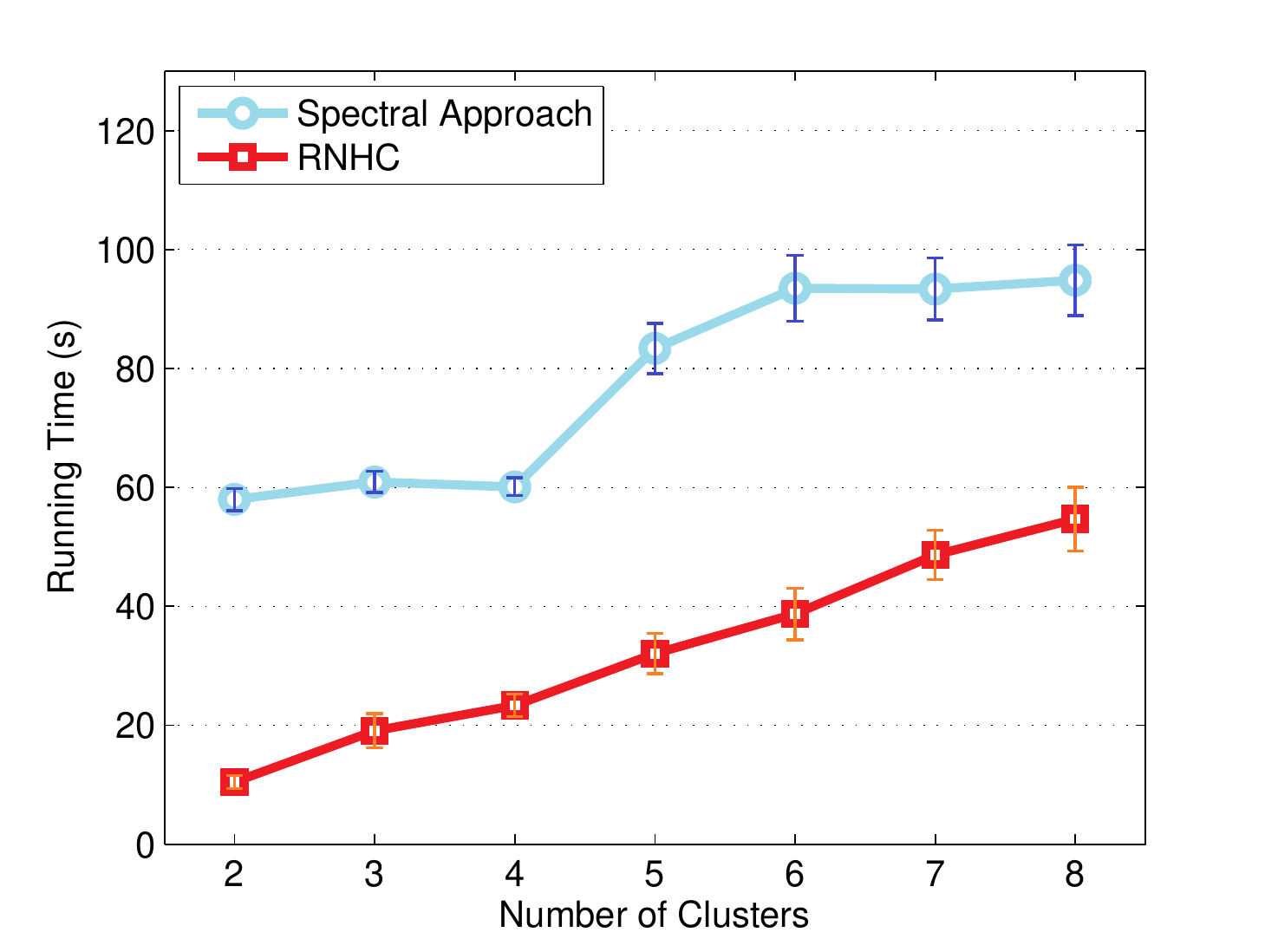}}
\subfigure[ibm16]{\includegraphics[width=0.22\textwidth,height=0.20\textwidth]{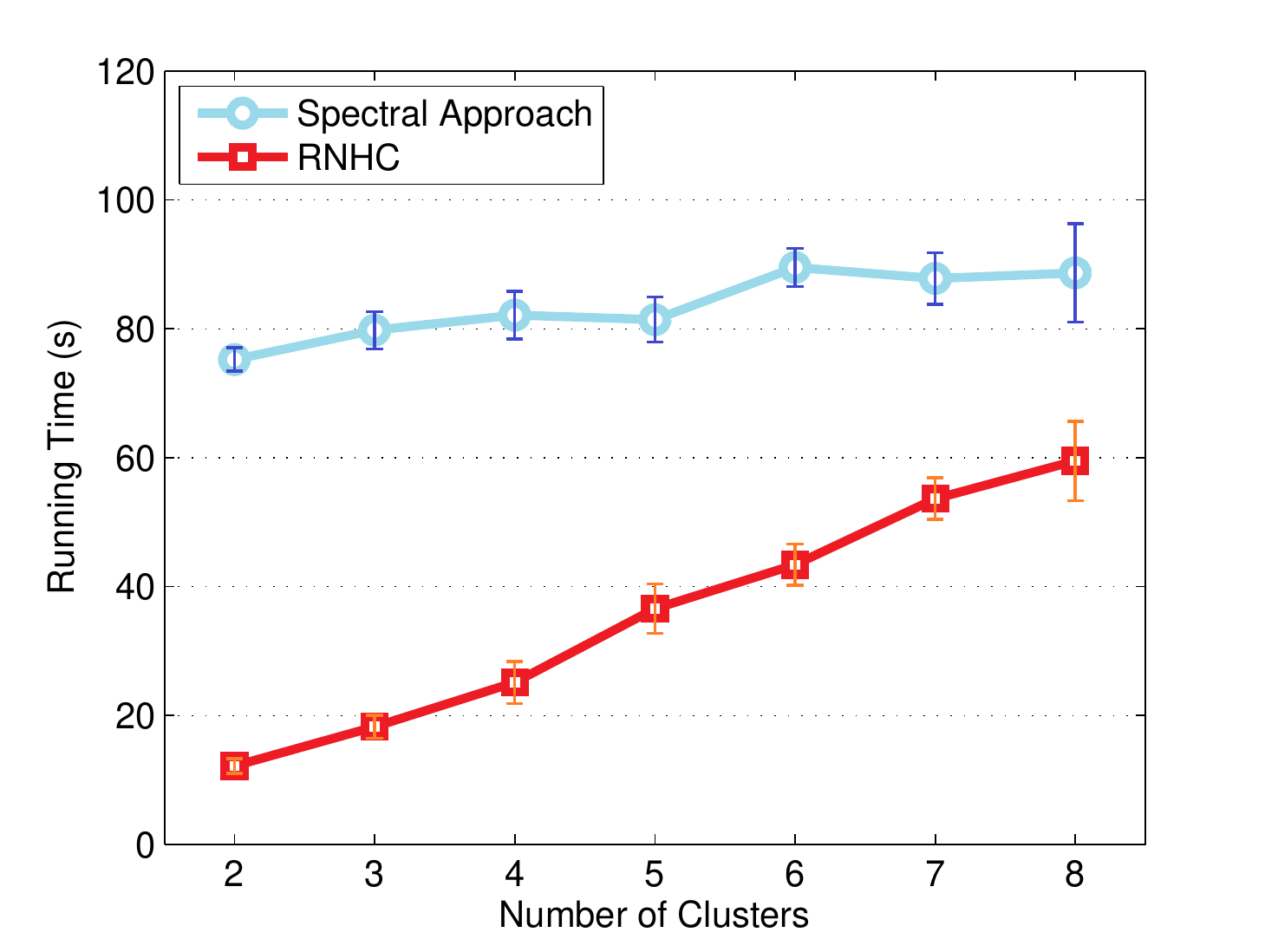}}
\subfigure[ibm17]{\includegraphics[width=0.22\textwidth,height=0.20\textwidth]{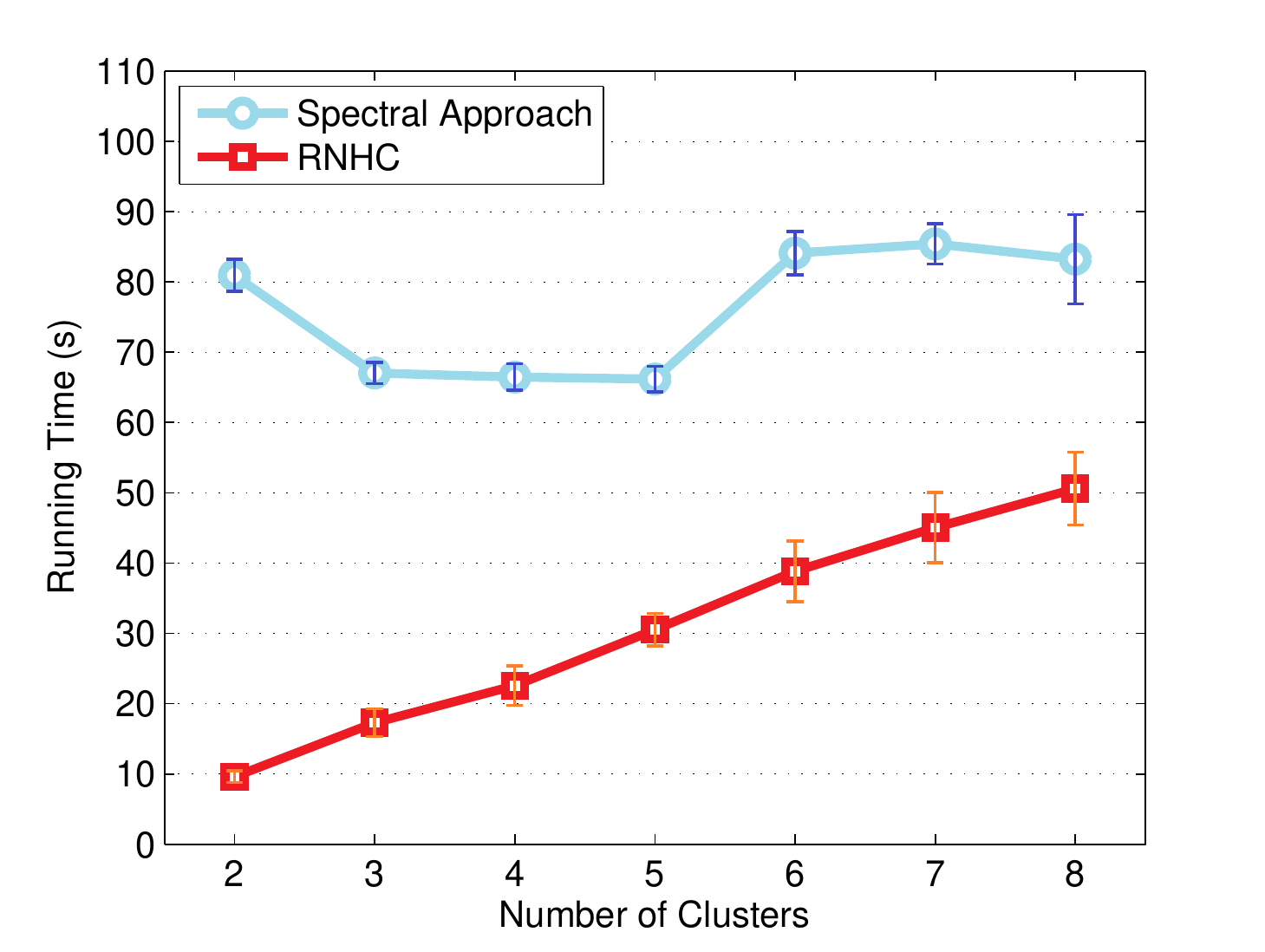}}
\subfigure[ibm18]{\includegraphics[width=0.22\textwidth,height=0.20\textwidth]{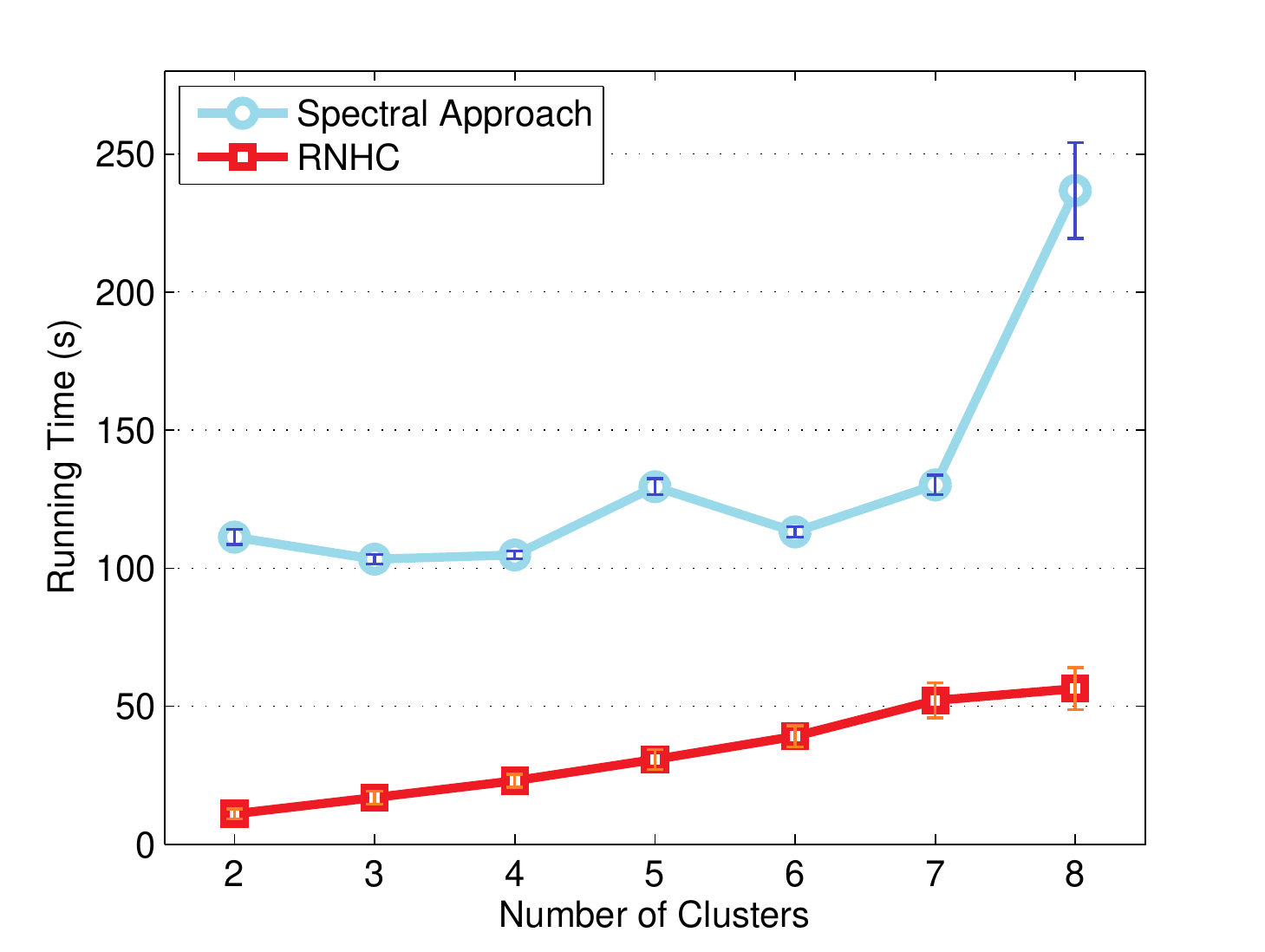}}
\caption{Running time of clustering on ibm15-18}
\label{fig:time}
\end{figure*}

The comparison of the time for clustering on the largest 4 datasets is shown in Figure~\ref{fig:time}. To guarantee fairness, all the experiments are carried out in a single thread by setting ``maxNumCompThreads(1)'' in MATLAB. We can find that our RNHC algorithm is faster than the baseline in all cases.

\section{Conclusion}
\label{section:conclusion}

In this paper we have proposed a new model to formulate the normalized hypergraph cut problem. 
Furthermore, we have   provided an effective  approach to relax  the new model, and
developed an efficient learning algorithm to solve the relaxed hypergraph cut problem. 
Experimental results on real hypergraphs have shown that our algorithm can outperform the state-of-the-art approach. 
It is interesting to apply our approach to other practical problems, such as the graph partitioning problem in distributed computation, in the future work.

\bibliographystyle{named}
\bibliography{RNHC}

\end{document}